%% file: main.tex
\documentclass{article} % For LaTeX2e
\usepackage{colla2022/collas2022_conference,times}
\collasfinalcopy

\input{math_commands.tex}

\usepackage{hyperref}
\hypersetup{
    colorlinks=true,
    linkcolor=red,
    filecolor=magenta,      
    urlcolor=blue,
    citecolor=purple,
    pdftitle={Overleaf Example},
    pdfpagemode=FullScreen,
    }

\usepackage{amsfonts}
\definecolor{lightgray}{gray}{0.9}
\usepackage{mathabx}
\usepackage{wasysym}
\usepackage{caption, subcaption}
\usepackage{wrapfig}
\usepackage{xparse}
\NewDocumentCommand{\codeword}{v}{%
\texttt{\small\textcolor{blue}{#1}}%
}

\usepackage[frozencache=true,cachedir=minted-cache]{minted}
\usepackage{listings}
\usepackage{bm}
\usepackage{graphicx}%,subcaption}4

\usepackage{pythonhighlight}
\usepackage{color, colortbl}
\definecolor{Gray}{gray}{0.9}

\usepackage{booktabs}

\usepackage{blindtext}

\def\listingautorefname{Listing}

\usepackage{etoolbox}
\makeatletter
\patchcmd{\hyper@makecurrent}{%
    \ifx\Hy@param\Hy@chapterstring
        \let\Hy@param\Hy@chapapp
    \fi
}{%
    \iftoggle{inappendix}{%true-branch
        \@checkappendixparam{chapter}%
        \@checkappendixparam{section}%
        \@checkappendixparam{subsection}%
        \@checkappendixparam{subsubsection}%
        \@checkappendixparam{paragraph}%
        \@checkappendixparam{subparagraph}%
    }{}%
}{}{\errmessage{failed to patch}}

\newcommand*{\@checkappendixparam}[1]{%
    \def\@checkappendixparamtmp{#1}%
    \ifx\Hy@param\@checkappendixparamtmp
        \let\Hy@param\Hy@appendixstring
    \fi
}
\makeatletter

\newtoggle{inappendix}
\togglefalse{inappendix}

\apptocmd{\appendix}{\toggletrue{inappendix}}{}{\errmessage{failed to patch}}
\usepackage[most]{tcolorbox}
\usepackage{todonotes}
\title{Sequoia: A Software Framework \\ to Unify Continual Learning Research}

\author{%
    Fabrice Normandin$^{123}$ \quad Florian Golemo$^{2}$ \quad Oleksiy Ostapenko$^{13}$ \quad Pau Rodr\'iguez$^2$ \\
    \bf Ryan Lindeborg  \quad Matthew Riemer$^{137}$ \quad Lucas Cecchi$^1$ \quad Timothée Lesort$^1$ \\ 
    \bf Khimya Khetarpal$^{14}$ \quad David Vazquez$^2$ \quad Laurent Charlin$^{156}$ \quad Irina Rish$^{136}$ \\
    \bf Massimo Caccia$^{123}$  \\
    \\
    $^1$Mila - Quebec AI Institute, $^2$ElementAI, a ServiceNow Company, $^3$Université de Montréal \\
    $^4$McGill University, $^5$HEC Montréal,  $^6$Canada CIFAR AI Chair, $^7$IBM Research\\
}

\begin{document}

\maketitle

\begin{abstract}
The field of Continual Learning (CL) seeks to develop algorithms that accumulate knowledge and skills over time through interaction with non-stationary environments. In practice, a plethora of evaluation procedures (\emph{settings}) and algorithmic solutions (\emph{methods}) exist, each with their own potentially disjoint set of \emph{assumptions}. This variety makes measuring progress in CL difficult. We propose a taxonomy of settings, where each setting is described as a set of assumptions. A tree-shaped hierarchy emerges from this view, where more general settings become the parents of those with more restrictive assumptions. This makes it possible to use inheritance to share and reuse research, as developing a method for a given setting also makes it directly applicable onto any of its children. We instantiate this idea as a publicly available software framework called \emph{Sequoia}, which features a wide variety of settings from both the Continual Supervised Learning (CSL) and Continual Reinforcement Learning (CRL) domains. Sequoia also includes a growing suite of methods that are easy to extend and customize, in addition to more specialized methods from external libraries. We hope that this new paradigm and its first implementation can help unify and accelerate research in CL. You can help us grow the tree by visiting \href{https://www.github.com/lebrice/Sequoia}{github.com/lebrice/Sequoia}.
\end{abstract}

%---------------------------------------------------------------------------------------------------
\section{Introduction}
\label{sec:introduction}

With the growing interest in developing methods robust to changes in the data distribution, research in continual learning (CL) has gained traction in recent years~\citep{delange2021continual, caccia2020online, parisi2019continual, lesort2019continual}. CL enables models to acquire knowledge from non-stationary data, learning new and possibly more complex tasks, while retaining performance on previously-learned tasks.

To instantiate a CL problem, one must first make assumptions about the data distribution and set constraints to enforce non-stationary learning. For instance, assumptions are often made about the type and number of tasks, the task boundaries, or the availability of task labels, while constraints often relate to memory, compute, or time.
Combinations of assumptions, rules, and datasets have resulted in a multitude of settings~\citep{khetarpal2020towards}.

We argue that the increased popularity of CL, combined with a lack of unification --- in part due to the large number of settings and the absence of well-defined applications --- has led to a ``research jungle'' that may be slowing down progress. 

% https://docs.google.com/drawings/d/1zs5fJfB0sWY4R8kDW-YIW1lpkVin2BG4Kq3eDPSiU7s/edit?usp=sharing
% https://docs.google.com/drawings/d/1sLU5XGOwFxi5BFkWfH9NvU0G2Y7AU-8x_1lLnJXU8oE/edit
\begin{figure}[t]
  \centering
        \includegraphics[width=0.8\linewidth, trim={0 0 0 2cm}]{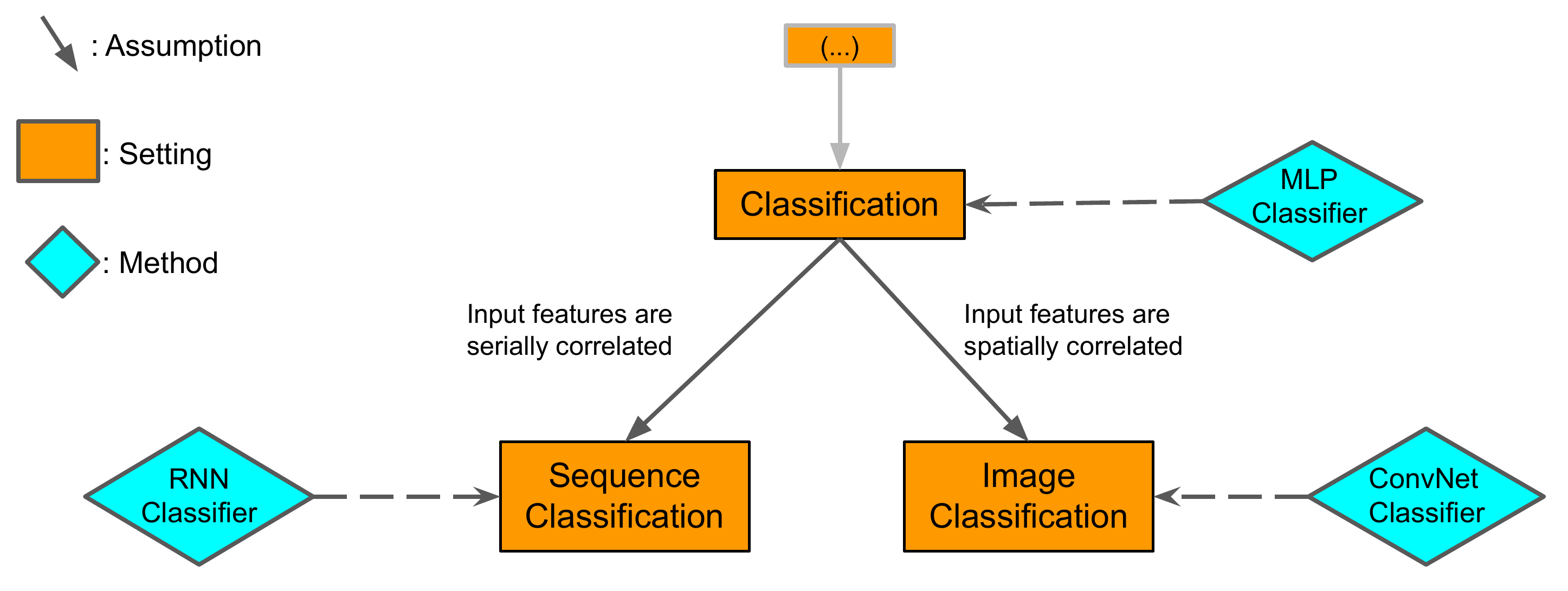} 
    \caption{Research settings can be organized into a hierarchy based on their assumptions. Methods have a \textit{target} setting (dashed arrows), and can be applied onto any of their descendants.
    In this trivial example, both MLP and ConvNet classifiers can be compared in the image classification setting, even though once makes more assumptions than the other. Sequoia applies this principle to the field of continual learning.
    }
    \label{fig:assumption_cartoon}
\end{figure}

We identify some of the main challenges associated with the lack of unification in CL:

\textbf{i) Evaluation.}
Methods in CL are often studied under a small subset of the available settings, making it difficult to evaluate them, as their problem domains don't always overlap. Consequently, it is challenging to determine if a method will generalize beyond the setting it was designed for. To add to this, continual reinforcement learning (CRL) poses further challenges in evaluation due to the lack of a clear distinction between training and testing phases~\citep{khetarpal2018re}. Moreover, resource consumption is a critical factor for evaluation of CL methods which is often overlooked due to the lack of standardized evaluation protocols. Thus, there is a need for standardization of the infrastructure used to evaluate CL methods.

\textbf{ii) Reproducibility.} In order to analyze specific properties of novel methods, researchers tend to re-implement baselines and adapt them to their particular needs~\citep{henderson2019deep}. These baselines are often not described in enough detail to ensure reproducibility, e.g.\ a prescribed hyper-parameter search strategy, computational requirements, open source libraries, etc.  

\textbf{iii) CSL and CRL evolve in silos.} Continual supervised learning (CSL) and continual reinforcement learning (CRL) are considered to be independent settings in the literature and thus, they are evolving separately. However, most methods in one field can be instantiated in the other, resulting in duplicate efforts such as replay for CSL~\citep{rebuffi2017icarl,lesort2018generative, Shin17, lesort2018marginal, prabhu12356gdumb} and replay for CRL~\citep{Traore19DisCoRL,rolnick2019experience, kaplanis2020continual}. To this end, we advocate that the unification of both fields would greatly reduce these duplicate efforts and accelerate CL research.

In this work, we present Sequoia, a unifying software framework for CL research, as a solution for jointly addressing these issues. 
We describe how \emph{settings} differ from one another in terms of their \emph{assumptions} (e.g., are task IDs observed or not). This perspective gives rise to a hierarchical organization of CL settings, through which methods become directly reusable by inheritance, thus greatly reducing the amount of work involved in developing and evaluating methods in CL. (See \autoref{fig:assumption_cartoon} for a cartoonish example).

% \textbf{Key Contributions} of this work are 1) a general unified framework that systematically organizes CL settings; 2) \emph{Sequoia}, a software framework that can serve as a universal platform for CL research.

\begin{tcolorbox}[enhanced,attach boxed title to top center={yshift=-3mm,yshifttext=-1mm},
  colback=blue!5!white,colframe=blue!75!black,colbacktitle=red!80!black,
  title=Key Contributions,fonttitle=\bfseries,
  boxed title style={size=small,colframe=red!50!black} ]
  1) a general unified framework that systematically organizes CL settings; \\
  2) \emph{Sequoia}, a software framework that can serve as a universal platform for CL research; \\
  3) an illustrative empirical study produced using Sequoia.
\end{tcolorbox}

%---------------------------------------------------------------------------------------------------

\section{A Unifying Framework for CL Research}
\label{sec:unify}

%  https://lucid.app/lucidchart/invitations/accept/inv_5f5f9a7d-8966-4e78-ae2a-aade6ffd7226
\begin{figure}[h]
  \centering
    \includegraphics[width=0.9\linewidth]{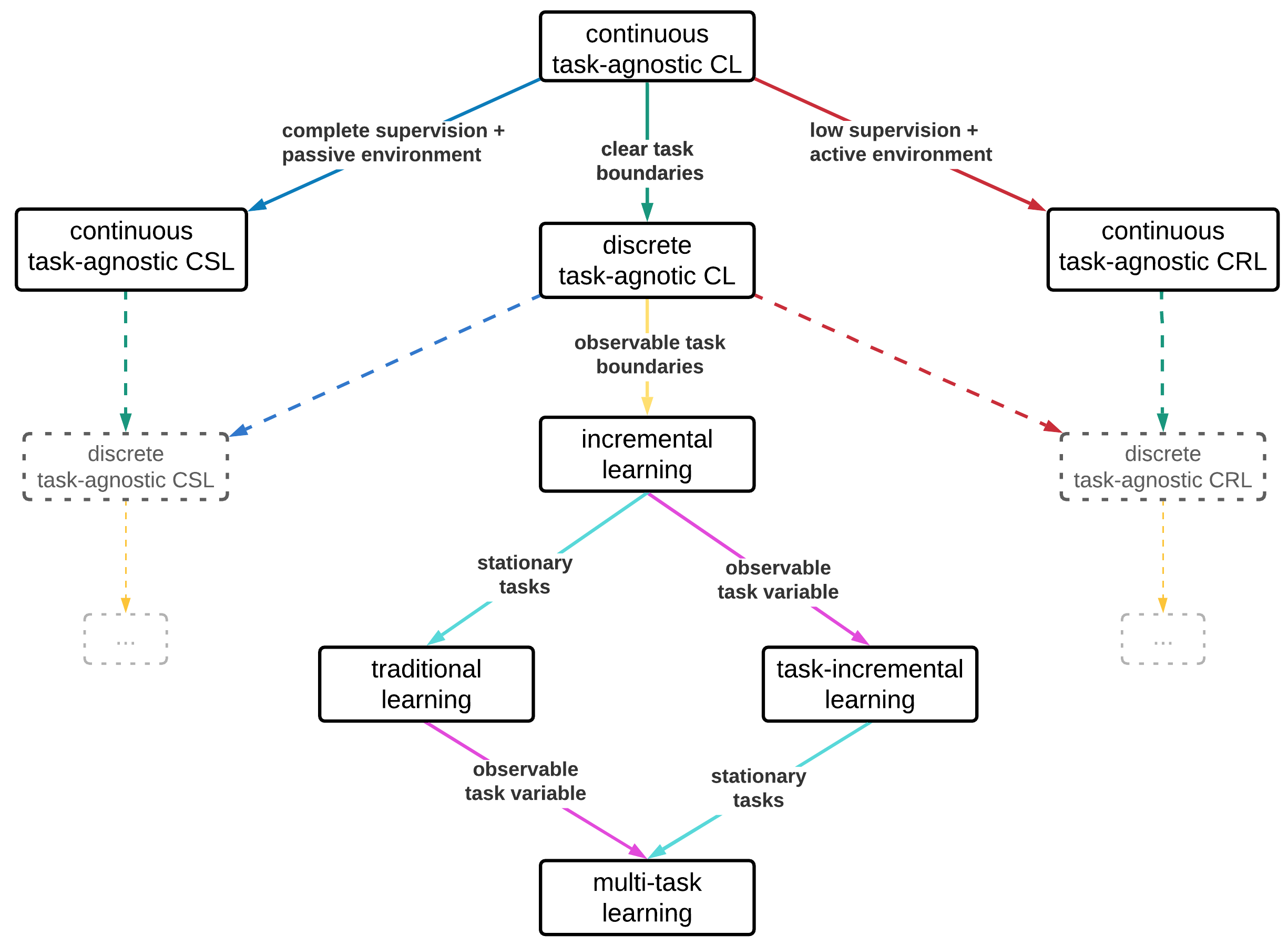} 
    \caption{\textbf{Sequoia - The Continual Learning Research Tree} Continual learning research settings can be organized into a tree, in which more general settings (parents) are linked with more restricted settings (children) by the differences in assumptions between them. Settings generally become more challenging the higher they are in this hierarchy, as less information becomes available to the method. The central portion of the tree shows the assumptions specific to CL, while the highest lateral branches indicate the choice of either supervised or reinforcement learning, which we consider to be orthogonal to CL. By combining either with the central assumptions, settings from Continual SL and Continual RL can be recovered to the left and right, respectively.}
    \label{fig:sequoia_cartoon}
\end{figure}
% \vspace{-0.5cm}

To construct our unifying framework, we first represent each CL setting as a set of shared assumptions. More general settings make fewer assumptions and vice versa. Settings can then be organized in a tree-shaped hierarchy\footnote{To be precise, the mathematical abstract structure of the framework is a lattice, and not a tree: the different settings form a \emph{partially} ordered set.} where adding/removing an assumption yields a child/parent setting  (\autoref{fig:sequoia_cartoon}). 

We formalize the framework using a generalization of the hidden-mode Markov decision process (HM-MDP), a special case of a POMDP~\citep{choi2000hidden}. HM-MDPs comprise an observation space $\mathcal{X}$, an action space $\mathcal{A}$, a context space $\mathcal{Z}$ (we also refer to contexts as tasks), and a feedback function $r$. Here the full state space is a concatenation of the observation space and task space $\mathcal{S}=\mathcal{X} \times \mathcal{Z}$. As such,  the hidden context variable  $\bm{z} \in \mathcal{Z}$ defines the dynamics of the environment $p(\bm{x}'|\bm{x},\bm{a},\bm{z})$ for observations $\bm{x}',\bm{x} \in \mathcal{X}$ and action $\bm{a} \in \mathcal{A}$. The feedback function $r(\bm{x},\bm{a},\bm{z})$ provides an agent $\pi(\bm{a}|\bm{x})$ (e.g. a supervised model) the value of performing particular actions $\bm{a}$ after receiving particular observations $\bm{x}$ (\autoref{sec:slrl} explains how the feedback functions returns targets in SL and rewards in RL). The context variable follows a Markov chain $p(\bm{z}'|\bm{z})$. The dynamic context enables modelling CL task/context-change. A change in the context variable is called a \emph{task boundary}. 

In the next section (\autoref{sec:cl_assumption}) we show that by restricting the different elements of the HM-MDP we recover different CL settings.  Then in \autoref{sec:slrl} we discuss differences between continual supervised (CSL) and continual reinforcement learning (CRL). We end in \autoref{sec:add_assumptions}, by presenting additional assumptions that are relevant to CL problems.

\subsection{Continual Learning Assumptions}
\label{sec:cl_assumption}

Assumptions related to CL can be arranged into a hierarchy, as illustrated in the the central portion of \autoref{fig:sequoia_cartoon}. These settings cover most, if not all the current CL literature. We start from the most general setting: continuous task-agnostic CL \citep{zeno2019task} and add assumptions one by one.

\textbf{Continuous Task-Agnostic CL} is our most general setting. The context variable is continuous $\mathcal{Z} \in \mathbb{R}$. This setting allows for different kinds of drifts in the environment, including smooth task boundaries, i.e. slow drift~\citep{zeno2019task}. This setting is \emph{task-agnostic}, meaning that the context variable $\bm{z}$ is unobserved.  Because the context is allowed to drift slowly, it can be more challenging for the methods to infer when a task has changed enough to compartmentalize the recently acquired knowledge before adapting to the new task. In RL, this setting is analogous to the DP-MDP~\citep{xie2020deep,chandakfuture}. In SL, it has also been studied in e.g. \citep{zeno2019task,Aljundi2019TaskFreeCL,Aljundi2019Gradient}. 

\textbf{Discrete Task-Agnostic CL} assumes clear (or well-defined) task boundaries and so a discrete context variable $\mathcal{Z} \in \mathbb{N}$. In this setting the context can shift in a drastic way, still without the agent being explicitly noticed. Some cases where this setting has been studied are ~\citet{choi2000hidden,riemer2018learning} for RL and ~\citet{caccia2020online,He2019TaskAC,Harrison2019ContinuousMW} for SL.  

\textbf{Incremental Learning} (IL) relaxes the task-agnostic assumption: the task boundaries are observable. This is akin to augmenting the observation with a binary variable that is set to 1 when $\bm{z}'\neq \bm{z}$ and 0 otherwise. In doing so, the algorithm does not need to perform \emph{task-boundary detection}. In SL, some well-known IL settings include class-IL and domain-IL distinguished by their \emph{disjoint action space} and \emph{shared action space}, respectively. This is discussed in \autoref{sec:add_assumptions}. 

At this point in the CL hierarchy, the tree branches in two directions, depending on the order of remaining assumptions (see \autoref{fig:sequoia_cartoon}). We will first explain the right sub-tree.

\textbf{Task-Incremental Learning} (task-IL) assumes a fully-observable context variable available to the agent $\pi(\bm{a}|\bm{x},\bm{z})$. In the literature, observing $\bm{z}$ is analogous to knowing the \emph{task ID} or \emph{task label}. In this simpler CL setting, forgetting can be prevented by freezing a model at the completion of each task and using the task-ID to retrieve it for evaluation.

The following settings remove the non-stationarity assumption in the contexts/tasks and are often used to set an upper-bound performance for CL methods.

\textbf{Multi-task Learning} removes the non-stationarity in the environment dynamics and the feedback function as it assumes a stationary context variable $p(\bm{z}'|\bm{z}) = p(\bm{z}')$. When the contexts are stationary, there is no \emph{catastrophic forgetting} (CF) \citep{French99} problem to solve. Multi-task learning assumes a fully-observable task variable.

\textbf{Traditional Learning} branches off incremental CL and assumes a stationary environment. It is the vanilla supervised setting machine learning defaults to. In our framework, it can be seen as a multi-task learning problem where the task variable isn't observable. However, a more natural view of this setting is to simply assume a single task/context.

\subsection{Supervised Learning and Reinforcement Learning Assumptions}
\label{sec:slrl}
So far we have introduced settings and assumptions that revolve mainly around the type and presence of non-stationarity in the environment and the information observed by the agent. These have allowed us to define the CL problem. To bring all of CL research under one umbrella, we introduce two assumptions, orthogonal to the previous ones, to recover RL and SL settings.
With these assumptions, methods for a given CL setting are applicable to both its CSL and CRL versions, as in ~\citet{kirkpatrick2017overcoming,fernando2017pathnet}.

Below we use the term observation as a \emph{state} in RL parlance and the actions as \emph{predictions} in SL parlance. Also, we assume a single context or task.

\textbf{Level of feedback:} In RL, the feedback function $r(\bm{x},\bm{a},\bm{z})$ returns a \emph{reward} that informs the agent about the value of performing action $\bm{a}$ after receiving observation $\bm{x}$ in context $\bm{z}$. In SL however, the feedback function is generally both directly known by the agent and differentiable, which  allows the agent to simultaneously consider the value of all actions for a particular observation. This feedback is computed based on a \emph{label} when the action space is discrete (classification) or a \emph{target} when it is continuous (regression). The feedback level is a key differentiating feature between RL and SL.

\textbf{Active vs passive environments:} In RL, it is generally assumed that the agent's action has an effect on the next observation or state.\footnote{The \emph{bandit} setting is one notable exception to this rule.}
In other words, the dynamics of the environment $p(\bm{x}'|\bm{x},\bm{a},\bm{z})$ are action-dependant and we call this an \emph{active} environment. In SL the agent is generally assumed to not influence the next observation  i.e. $p(\bm{x}'|\bm{x},\bm{a},\bm{z})=p(\bm{x}'|\bm{x},\bm{z})$. The environment is thus referred to as being \emph{passive} in these cases.

As seen in \autoref{fig:sequoia_cartoon}, the two aforementioned assumptions are combined into a single assumption for SL (blue, left) and for RL (red, right). By combining either the RL or SL assumption along with those from the the central CL ``trunk'', settings from  CSL and CRL are recovered. Future versions of Sequoia will decouple these assumptions to enable settings such as bandits and imitation learning.

\subsection{Additional Assumptions}
\label{sec:add_assumptions}

Additional assumptions can be added on top of the ones described above to recover additional research settings.
For example, a useful assumption in CL experiments is the one of disjoint versus joint action space, i.e. whether the contexts/tasks share a same action space, or whether that space is different for each task. In CSL, this assumption differentiates \emph{class-incremental learning} from \emph{domain-incremental learning} \citep{van2019three}. In~\citet{Farquhar18}, where it is referred as the \emph{shared output space} assumption, a disjoint action space greatly increases the difficulty of a setting in terms of forgetting. In CRL however, the studied settings mostly have a joint action space, with the notable exception in the work of ~\citet{chandak2020lifelong}.

Other assumptions could also be relevant in defining a continual learning problem. For instance, the action space being either discrete or continuous, resulting in classification and regression CSL problems, respectively; a particular structure being required of the method's actions, as in image segmentation problems; an episodic vs non-episodic setting in RL; context-dependant \citep{caccia2020online} versus context-independent feedback functions; and many more.

%---------------------------------------------------------------------------------------------------
\section{Sequoia - A Software Framework}

Alongside this unifying perspective, we introduce \emph{Sequoia}, an open-source python framework. Each setting described above is instantiated as a class in a tree-shape inheritance hierarchy.
Sequoia is designed to address some of the issues associated with Continual Learning research, previously described in \autoref{sec:introduction}.

% Method evaluation / Reproducibility / Baseline re-implementation  / CRL vs CRL evolve in silos.

First, we establish a clear \textbf{separation of concerns} between research problems and the solutions to such problems. We establish this separation through two core abstractions: \codeword{Setting} and \codeword{Method}. This decoupling greatly helps to evaluate methods, since the logic for each component is cleanly separated, and extracting a component and reusing it elsewhere becomes possible. An example of a Method is shown in ~\autoref{fig:minimal_method}.

Second, to help bridge the gap between the CRL and CSL domains, Sequoia uses \codeword{Environment} as the interface between methods and settings. This class extends the familiar \codeword{Env} abstraction from OpenAI \codeword{gym} to also include supervised learning datasets, making it possible to develop methods that are applicable in both the CRL and CSL domains. \codeword{Environment} will be described in \autoref{sec:environments}.

Finally, Sequoia uses inheritance to \textbf{make methods directly reusable across settings}. By organizing research settings into a tree-shaped inheritance hierarchy, along with their environments, observations, actions, and rewards, Sequoia enables methods developed for any particular setting to be applicable onto any of their descendants, since all the objects the method will interact with will inherit from those they were designed to handle. This mechanism has the potential to greatly improve code reuse and reproducibility in CL research.

This section first describes each of these abstractions in more detail, after which the currently available settings and methods are described. \autoref{sec:experiments} will then provide a demonstration of the kind of large-scale empirical studies which are made possible through the use of this new framework.

\begin{wrapfigure}[20]{r}{0.42\textwidth}
% \begin{figure}[h]
% \begin{figure}[t]
    \centering
        \scriptsize
        % \inputminted[hoptionsi]{hlanguage}{filename}
\begin{minted}{python}
from sequoia.settings.sl import *
from sequoia.settings.rl import *
from sequoia.methods import BaseMethod

method = BaseMethod(learning_rate=1e-3)

for setting in [
    ContinuousTaskAgnosticSLSetting("mnist"),
    ContinuousTaskAgnosticRLSetting("cartpole"),
    DiscreteTaskAgnosticSLSetting("mnist"),
    DiscreteTaskAgnosticRLSetting("cartpole"),
    IncrementalSLSetting("mnist"),
    IncrementalRLSetting("cartpole"),
    TaskIncrementalSLSetting("mnist"),
    TaskIncrementalRLSetting("cartpole"),
    MultiTaskSLSetting("mnist"),
    MultiTaskRLSetting("cartpole"),
    TraditionalSLSetting("mnist"),
    TraditionalRLSetting("cartpole"),
]:
    results = setting.apply(method)
    results.summary()
    results.make_plots()
\end{minted}
% \lstinputlisting[frame=single,breakindent=.5\textwidth,frame=single,breaklines=true,style=mypython,basicstyle=\small]{listings/run_multiple_settings.py}
        \captionof{listing}{Code snippet, using Sequoia to \\ evaluate a method in multiple settings.}
        \label{fig:multiple_settings}
% \end{figure}
\end{wrapfigure}

\textbf{Relation with other frameworks:} 
Sequoia is in no way competing with existing tools and libraries which provide standardized benchmarks, models, or algorithm implementations.
On the contrary, Sequoia benefits from the development of such frameworks.

In the case of libraries that introduce standardized benchmarks, they can be used to enrich existing settings with additional datasets or environments \citep{douillard2021continuum,brockman2016openai}, or even to create entirely new settings.
Likewise, frameworks which introduce new models or algorithms  \citep{lomonaco2021avalanche,stable-baselines3,ContinualWorld} can also be used to create new Methods or to add new backbones to existing Methods within Sequoia.
External repositories can register their own methods through a simple plugin system.
% Sequoia's main objective is to organize and structure research problems in CL in a way that makes solutions more easily reusable and reproducible.
The end goal for Sequoia is to provide the research community with a centralized catalog of the different research frameworks and their associated methods, settings, environments, etc.
The following sections will show examples of such extensions.

% This section describes these aspects of Sequoia, along with a brief description of the settings and methods currently available. The following section will then provide a glimpse of the kind of large-scale empirical studies which are made possible through this new framework. 

\subsection{Settings}
\label{subsec:settings}
A \codeword{Setting} can be viewed as a configurable evaluation procedure for a \codeword{Method}. %, that accepts a \codeword{Method} as input
It creates various training environments, evaluates the method, and finally returns some \codeword{Results}. These results contain various metrics relevant to the setting.
The training/testing routine for each setting is implemented according to the evaluation protocol of that setting in the literature.
An example of applying a method onto multiple settings is shown in ~\autoref{fig:multiple_settings}.

Concretely, settings create the training, validation, and testing environments that a method interacts with. This interface also makes it possible for methods to leverage PyTorch-Lightning~\citep{falcon2019pytorch} to perform high-performance training of their models.\footnote{It is important to note that methods are in no way required to use PyTorch-Lightning.}
See \autoref{app:adding_new_settings} for more details on the interactions between Sequoia and PyTorch-Lightning.

Settings are available for each combination of the CL assumptions, along with the choice of one of RL / SL (as illustrated in \autoref{fig:sequoia_cartoon}), for a total of 12 settings.\footnote{Other common SL settings, such as Domain-Incremental learning are also available in Sequoia, but they rely on an additional family of assumption, and are thus omitted from the main portion of this paper for sake of brevity and clarity.}
These two ``branches'' (one for CRL and the other CSL) form the basis of Sequoia's eponymous tree of settings.
Each setting inherits from one or more parent settings, following the above-mentioned organization.

Every \codeword{Setting} is created by extending a more general \codeword{Setting} and adding additional assumptions.
This inheritance relationship from one setting to the next also extends to the setting's environments (\codeword{Environment}), as well as the objects (\codeword{Observations}, \codeword{Actions}, and \codeword{Rewards}) they create.
See \autoref{app:settings_uml} for an illustration of this principle.

\subsection{Environments}
\label{sec:environments}
Settings in Sequoia create training, validation, and testing \codeword{Environment}s, which adhere to both the \texttt{gym.Env} and the \texttt{torch.DataLoader} APIs. This makes it easy for SL researchers to transition to RL and vice-versa.
These environments receive \codeword{Actions} and return \codeword{Observations} and \codeword{Rewards}. \codeword{Observations} contain the input samples \codeword{x}, and may also contain task labels for each sample, depending on the setting.
These objects have the same structure in both RL and SL settings. However, as described in \autoref{sec:slrl}, in SL, \codeword{Actions} correspond to the predictions, while \codeword{Rewards} correspond to targets or labels.
These objects are defined by the \codeword{Setting} and follow the same pattern of inheritance as the settings themselves. 
The structure of these objects are reflected in the environment's observation, action, and reward spaces, which are used within methods to create their models.

\paragraph{Supervised Learning environments}

Sequoia supports most of the datasets traditionally used in continual supervised learning research, through its use of the \textit{Continuum} package~\citep{douillard2021continuum}. The benchmarks from the CTrL package~\cite{ctrl} are also available as an optional extension. The list of supported datasets is available in \autoref{tab:support}. 

\paragraph{Reinforcement Learning environments}

Through its close integration with \codeword{gym}, Sequoia is able to use any gym-compatible environment as the ``dataset'' used by its RL settings.
For settings with multiple tasks, Sequoia simply needs a way to sample new tasks for the chosen environment. This mechanism makes it easy to add support for new or existing gym environments. An example of this is included in \autoref{app:add_new_env}.

Sequoia creates continuous or discrete tasks, depending on the choice of setting and dataset/environment.
For example, when using one of the classic-control environments from gym such as \texttt{CartPole}, tasks are created by sampling a new set of values for the environment constants such as the gravity, the length of the pole, the mass of the cart, etc.
This is also the case for the well-known \texttt{HalfCheetah}, \texttt{Walker2d}, and \texttt{Hopper} MuJoCo environments, where tasks can be created by introducing changes in the environmental constants such as gravity.
Continuous tasks can thus easily be created in this case, as the environment is able to respond dynamically to changes in these values at every step, and the task can evolve smoothly by interpolating between different target values.

Other gym environments become available when using RL settings with discrete tasks (i.e. all settings that inherit from \codeword{DiscreteTaskAgnosticRLSetting}), as it becomes possible to simply give Sequoia a list of environments to use for each task, and the constructed Setting will then use them as part of its evaluation procedure.

We use this feature to construct continual variants of the MT10, MT50 benchmarks from Meta-World~\citep{MetaWorld}, as well to replicate the CW10 and CW20 benchmarks introduced in \citep{ContinualWorld}.
Sequoia also introduces a non-stationary version of the \textit{MonsterKong} environment, described in  ~\autoref{app:continual_monsterkong}.
A more complete list of the supported environments is shown in \autoref{tab:support}.

% A growing number of libraries and tools have been recently introduced in the realm of few-shot/Meta-RL, (Meta-World, procgen, etc), which can easily be used to create such tasks.

\newcommand{\Empty}{$\emptyset$}
\newcommand{\Compiles}{$\sim$}
\newcommand{\AdaptedFor}{$\checkmark$}
\newcommand{\TableColWidth}{0.85cm}

%% here: https://docs.google.com/spreadsheets/d/1QR_UPlpLwokRzZCAq0MYDsiTn3WyXKJguYgyItU39Eo/

\begin{table*}
    \centering
    \begin{footnotesize}
    \begin{tabular}{l|p{0.02\linewidth} p{0.805\linewidth}}
        %  \rule{0pt}{4ex}
         Methods

         % empty
         & SL &
         \codeword{BaseMethod}.\{base, EWC, PackNet \}, 
         PNN, replay, HAT, CN-DPM
        \newline
        \codeword{Avalanche}.\{naive, AGEM, CWR$^*$, EWC, Gdumb, GEM, LWF, replay, SI\}
        % \newline
         \\
         
         & RL &
         \codeword{BaseMethod}.\{base, EWC, PackNet \}, 
        %  La-MAML \citep{gupta2020maml},
         PNN
         \newline
         \codeword{stable_baselines3}.\{A2C, DDPG, DQN, PPO, SAC, TD3\}
         \newline
         \codeword{continual_world}.\{SAC, AGEM, EWC, VCL, PackNet, L2 reg., MAS, replay\}
        %  \vspace{0.1cm}
         \\

         \hline
        %  \vspace{0.2c}
         \rule{0pt}{2ex}Environments & SL &
        %  \rule{0pt}{4ex}Environments & SL &
        %  \codeword{continuum}.\{\{K,E,Q,Fashion\}MNIST, 
        %  Cifar10(0),
        %  ImageNet100(0), 
        %  Core50,
        %  Synbols\}%, CUB200, AwA2, TinyImageNet200\} % TODO: Add these once they are truly supported.
        \codeword{continuum}.\{\{K,E,Q,Fashion\}MNIST, 
         Cifar10(0),
         ImageNet100(0), 
         Core50,
         Synbols\}%, CUB200, AwA2, TinyImageNet200\} % TODO: Add these once they are truly supported.
        \newline
        \codeword{CTrL}.\{s\_plus, s\_minus, s\_in, s\_out, s\_pl\}
        %  \vspace{0.1cm}
         \\
         
        %  \hline
         
        %  \rule{0pt}{3ex} Environments
         & RL &
         \codeword{gym}.\{Hopper, Half-Chettah,
         Walker2d, CartPole, Pendulum, MontainCar\}
         Monsterkong,
         \newline
         \codeword{metaworld}.\{MT10, MT50\}, \codeword{continual_world}.\{CW10, CW20\}
        %  \newline
        %  \vspace{0.cm}
         \\
         \hline
        %  \rule{0pt}{3ex} Metrics & &
        %  \{Transfer Matrix (forward \& backward transfer), Average final performance, Online Training Performance\}
        %  $\times$
        % %  \newline
        %  \vspace{0.2cm} \\
         \rule{0pt}{2ex}Metrics & & 
         \{Transfer Matrix, forward transfer, backward transfer, Average final performance, \newline Online Training Performance\}
         $\times$
         \\
         & SL & 
         \{loss, accuracy\}
         \vspace{0.1cm} \\
         & RL &
         \{loss, total reward, average reward, episode length\}
        \\
    \end{tabular}
    \caption{\textbf{Sequoia's methods, environments and metrics.} Most of the RL settings in Sequoia can be passed custom environments to use for each task. This makes it possible to use virtually any gym environment to create custom incremental RL settings. The environments listed here are those explicitly supported in Sequoia, where multiple tasks can be sampled within a single environment.
    }
    \label{tab:support}
    \end{footnotesize}
\end{table*}

\subsection{Methods}
\label{subsec:methods}

Methods hold the logic related to the model and training algorithm. When defined, each method selects a ``target setting'' from those available in the tree. A method can be applied to its target setting as well as any setting which inherits from it (i.e. any setting which is a child node of the target setting).
We now provide a brief description of the different types of Methods available in Sequoia. An illustration of the Method API can be seen in ~\autoref{fig:minimal_method}.

\begin{listing}[t]
    \scriptsize
    % Input the file directly. Makes this a bit easier to read.
\begin{minted}{python}
import gym
from sequoia.settings import Setting, Environment, Observations, Actions
from sequoia.methods import Method

class DemoModel:
    def forward(self, observations: Observations) -> Actions:
        ...

class DemoMethod(Method, target_setting=Setting):
    """ Pseudocode for a Method that targets a given Setting. """  
    def configure(self, setting: Setting):
        # Called by the setting before training begins.
        self.model = DemoModel(setting.observation_space, setting.action_space,
                               setting.reward_space, nb_tasks=setting.nb_tasks)
        self.optimizer = Adam(...)

    def fit(self, train_env: Environment, valid_env: Environment):
        # Train a model using these environments from the setting.
        # Note: all Environments are gym environments. More on this later.
        for epoch in range(self.n_epochs):
            self.model.train_epoch(train_env)
            self.model.validation_epoch(valid_env)

    def get_actions(self, observations: Observations) -> Actions:
        # Called by the setting for inference (at test-time).
        actions = self.model(observations)
        return actions

    def on_task_switch(self, task_id: Optional[int]):
        # Gets called on task boundaries, depending on the setting.
        self.model.prepare_for_new_task(new_task=task_id)
\end{minted}
    % \lstinputlisting[frame=single,breakindent=.5\textwidth,frame=single,breaklines=true,style=mypython]{listings/minimal_method.py}
    \caption{Pseudocode for creating a new Method in Sequoia.
    % Methods in Sequoia need to implement \pyth{fit}, which is used for training and validation, as well as \pyth{get_actions}, which is used for inference (at test-time). Both \texttt{configure} and \texttt{on\_task\_switch} are optional hooks: \texttt{configure} is called before training so that the method can adapt itself to the Setting, and \texttt{on\_task\_switch} may be called when a task boundary is reached, and may be passed the id of the new task, depending on the chosen setting.
    % Also note that the \texttt{Environment} class extends the gym.Env 
    % See \autoref{app:methods} for a more detailed description of the Method API.
    }
    \label{fig:minimal_method}
\end{listing}
%% Attempt at a "narrower" figure that we could wrap around.
% \begin{listing}[t]
%     \scriptsize
%     % Input the file directly. Makes this a bit easier to read.
%     \inputminted[]{python}{listings/minimal_method_narrow.py}
%     % \lstinputlisting[frame=single,breakindent=.5\textwidth,frame=single,breaklines=true,style=mypython]{listings/minimal_method.py}
%     \caption{Pseudocode for creating a new Method.
%     % Also note that the \texttt{Environment} class extends the gym.Env 
%     % See \autoref{app:methods} for a more detailed description of the Method API.
%     }
%     \label{fig:minimal_method}
% \end{listing}

\textbf{General methods: }
% TODO: Describe/explain that Sequoia comes with some built-in methods, but also enlist the help of 
Methods in Sequoia can target settings from either the RL or SL branches of the tree. Additionally, it is also possible to select one of the settings from the central CL branch - for instance, Incremental Learning. This makes methods applicable to both the CRL and CSL variants of that setting.
One such method is the \codeword{BaseMethod}, which can be applied to any setting in the tree, and is provided as a modular, customizable, jumping off point for new users. This \codeword{BaseMethod} is equipped with modules for task inference and multi-head prediction. See \autoref{app:base_method} for a more in-depth discussion of its features and capabilities.

\textbf{Supervised Learning Methods: }
Sequoia benefits from other CL frameworks such as Avalanche \citep{lomonaco2021avalanche}. Avalanche offers both standardized benchmarks as well as a growing set of CL methods, which are referred to as \textit{strategies} in Avalanche. Sequoia reuses these strategies as \codeword{Method} classes. See \autoref{tab:support} for a complete list of such methods.

\textbf{Reinforcement Learning Methods: }
Settings in Sequoia produce Environments, which adhere exactly to the Gym API. It is therefore easy to import existing RL tools and libraries and use them to create new methods. 

As an example, here we enlist the help of a specialized framework for RL, namely stable-baselines3~\citep{stable-baselines3} . The A2C, PPO, DQN, DDPG, TD3 and SAC algorithms from SB3 were easily introduced as new \codeword{Method} classes in Sequoia, without duplicating any code. These methods are applicable onto any of the RL settings in the tree.

In addition to these RL backbones from SB3, CRL methods are also available. These methods were adapted from the work of~\citet{ContinualWorld}, which introduced the CW10 and CW20 benchmarks for continual learning, based on sequences of tasks from Meta-World~\citep{MetaWorld}. The authors also provided implementations for CRL algorithms, built on top of a SAC~\citep{haarnoja2018soft} backbone. These algorithms were adapted from their original implementation and made available as CRL methods in Sequoia (see \autoref{tab:support} for full the list).\footnote{While most other methods use PyTorch these methods are implemented using Tensorflow.} 
% These methods can be seen in \autoref{tab:support}.

%---------------------------------------------------------------------------------------------------
\section{Experiments}
\label{sec:experiments}

Sequoia's design makes it easy to conduct large-scale experiments to compare the performance of different methods on a given setting, or to evaluate the performance of a given method across multiple settings and datasets.
We illustrate this by performing large-scale empirical studies involving all the settings and methods available in Sequoia, both in CRL and CSL.
Each study involves up to 20 hyper-parameter configurations for each combination of setting, method, and dataset, in both RL and SL, for a combined total of $\approx8000$ individual completed runs.
The rest of this section provides a brief overview of these experiments, which are also publicly available at \url{https://wandb.ai/sequoia/}.\footnote{We will update these sample studies periodically to reflect all future improvements made to the framework.} All results are reproduced in \autoref{app:ext_exp} in a larger format and accompanied with further analysis.

\begin{figure}
  \centering
%   \vspace{-3mm}
    \includegraphics[width=0.262\linewidth]{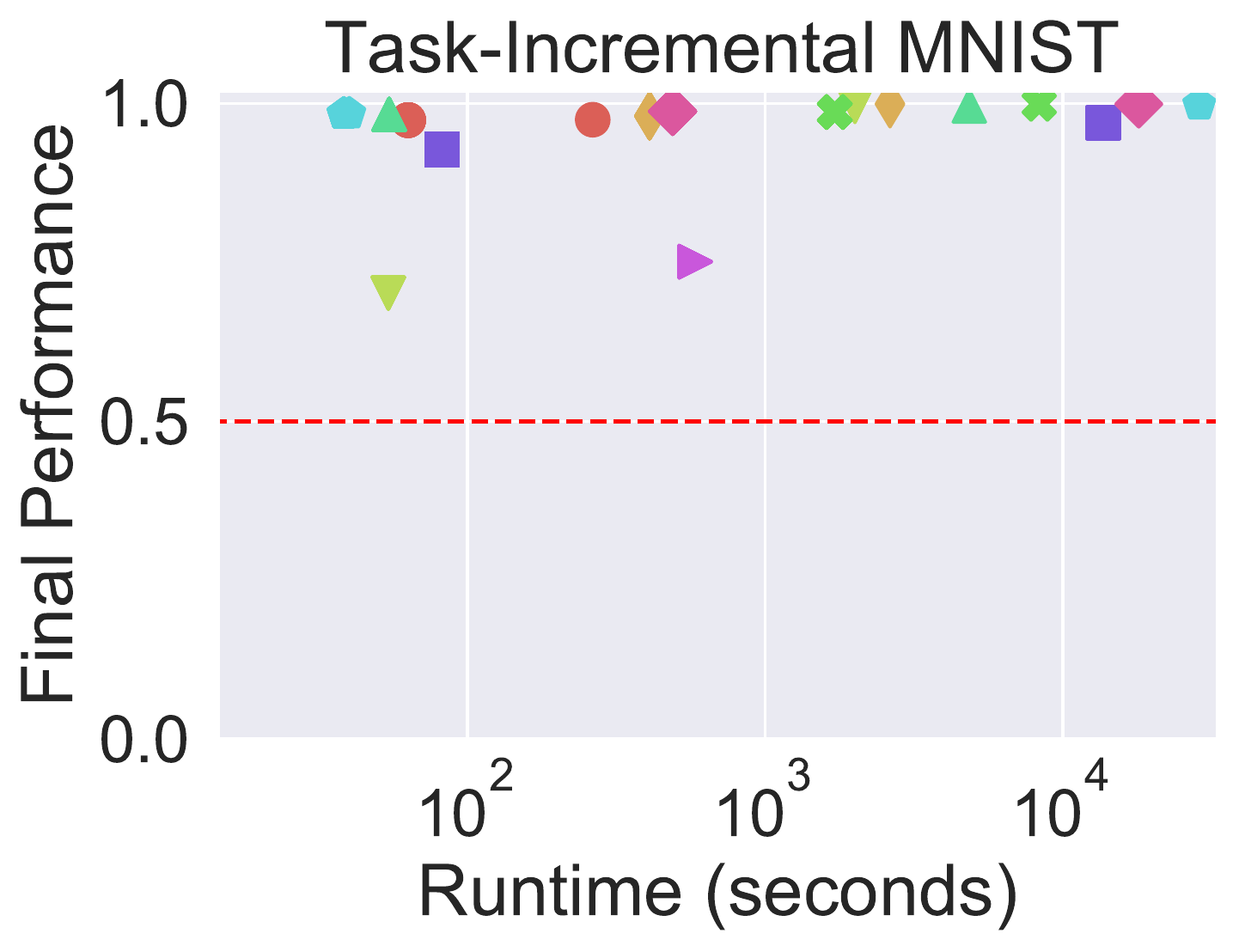} 
    \includegraphics[width=0.262\linewidth]{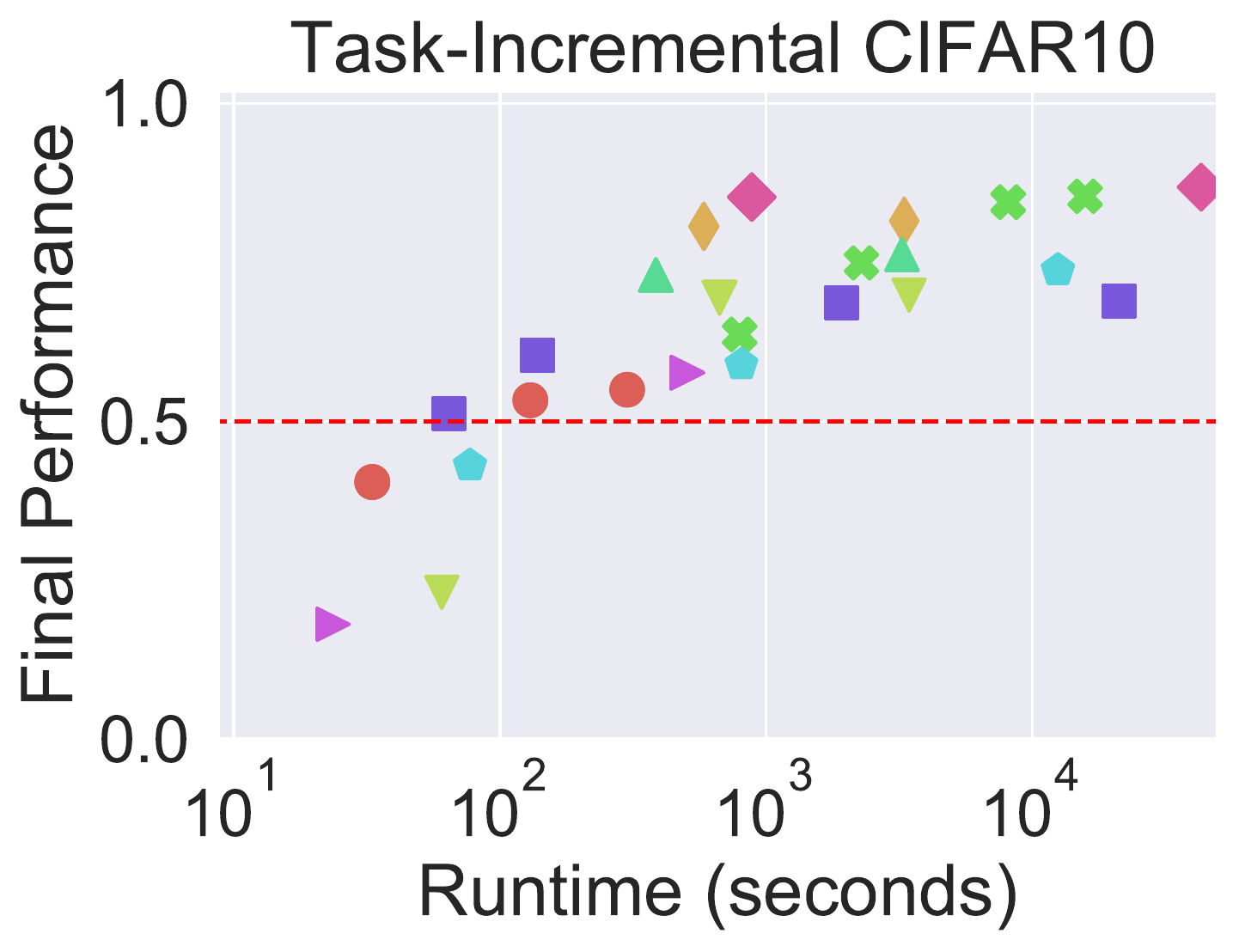} 
    \includegraphics[width=0.442\linewidth]{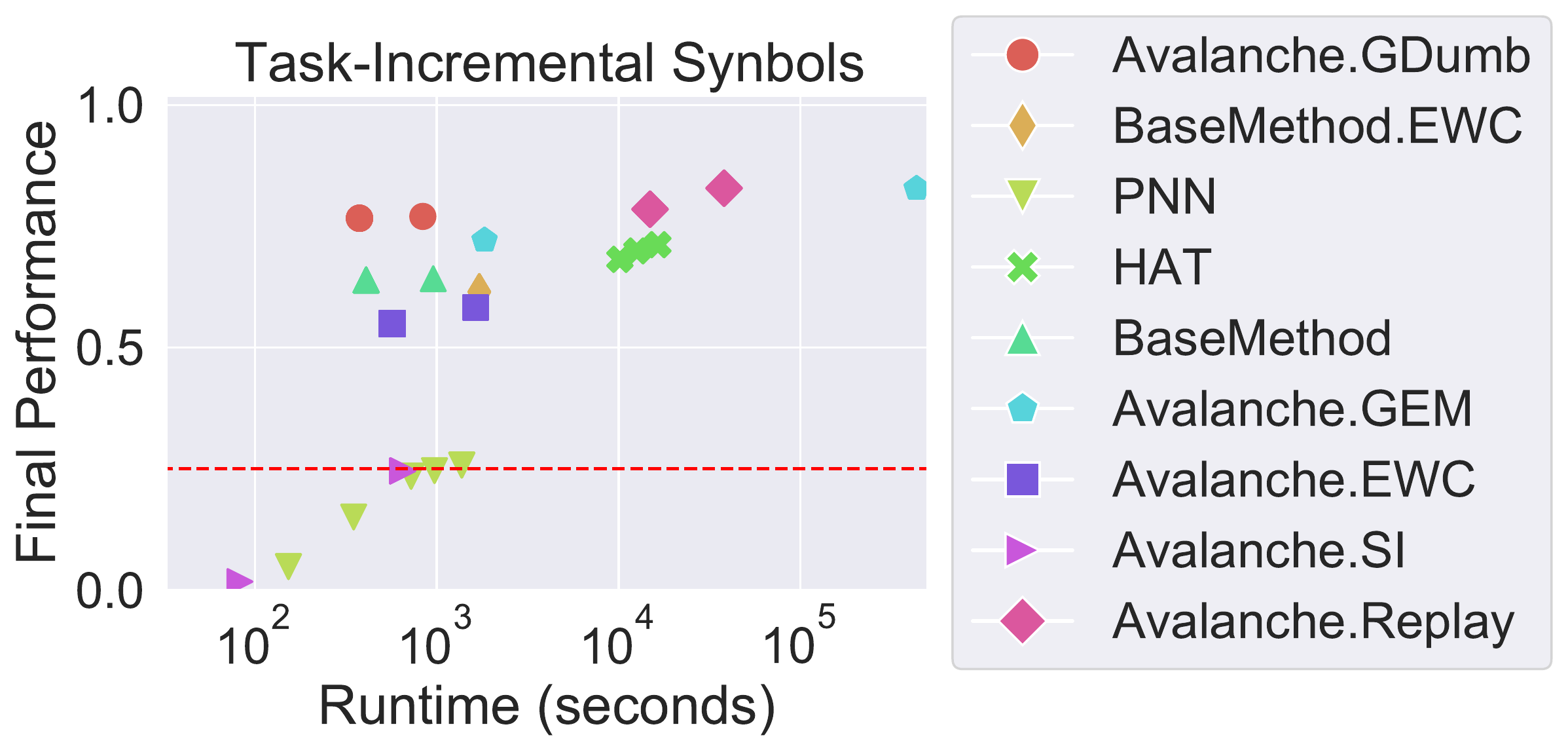} \\
    \includegraphics[width=0.262\linewidth]{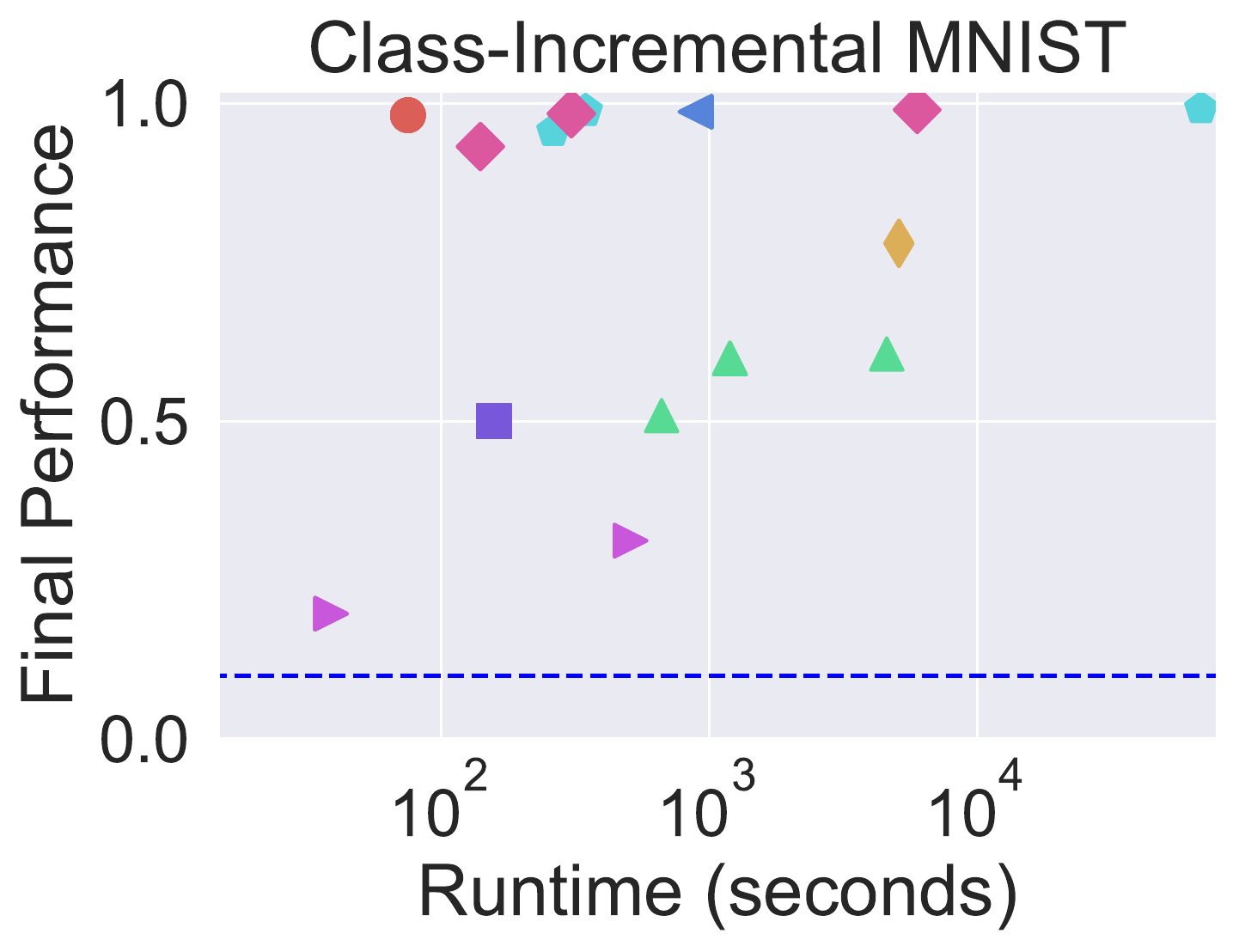} 
    \includegraphics[width=0.262\linewidth]{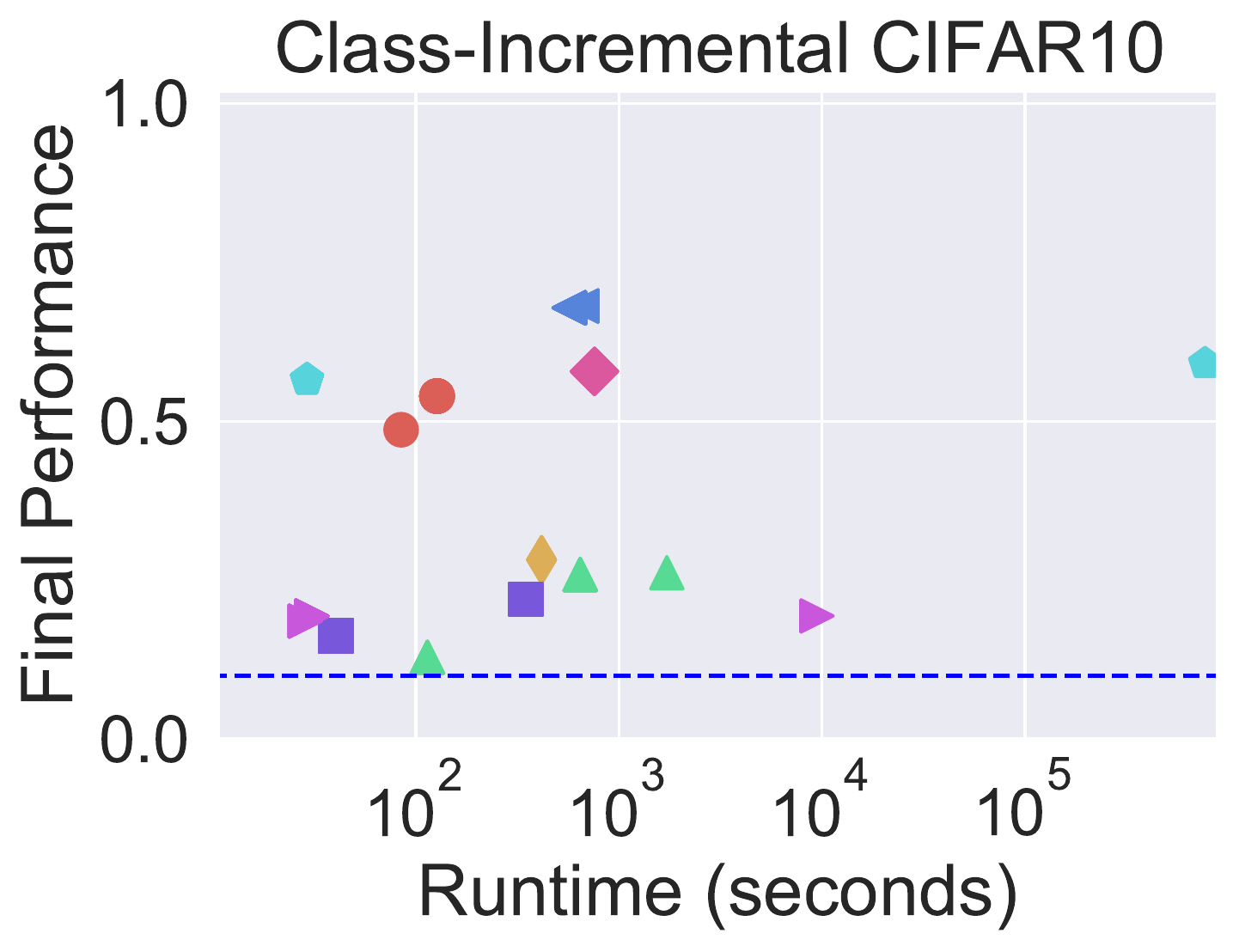} 
    \includegraphics[width=0.442\linewidth]{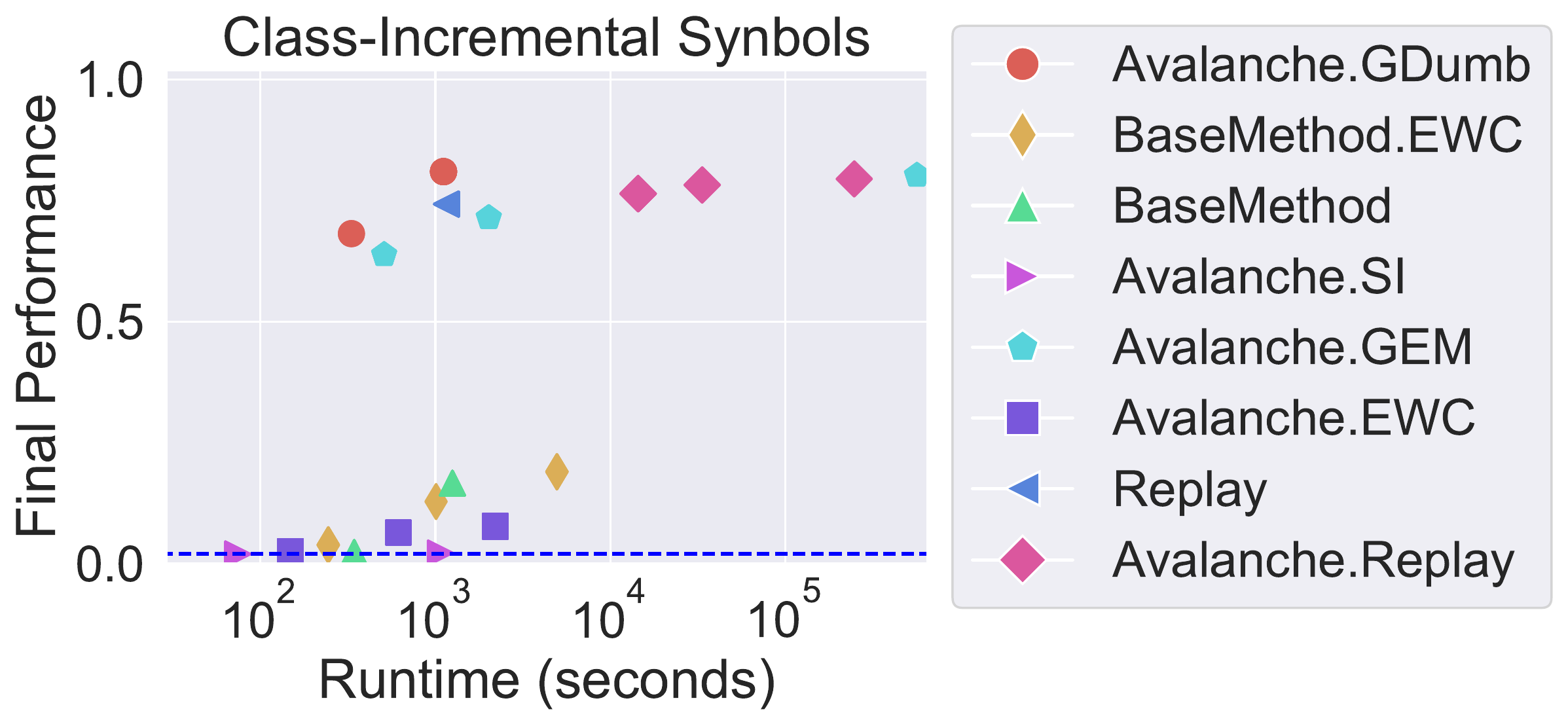} \\
    \caption{\textbf{Incremental Supervised Learning results}. Final performance (vertical axis) is plotted against runtime (horizontal axis). The methods achieving the best trade-off lie closer to the top-left of the figures. Task-Incremental and Class-Incremental results are presented on the top and bottom row, respectively. The dotted line shows chance accuracy for each setting-dataset combination. For each methods, several trials are presented depending on metrics composed of linear combination of final performance and (normalized) runtime. GEM and GDumb achieve the best tradoff, although the latter cannot make predictions in an online manner and thus serves more as a reference point. }
    \label{fig:incremental_sl}
    % \vspace{-3mm}
\end{figure}

%------------------------
\textbf{Continual Supervised Learning.} As part of the CSL study, we use some of the ``standard'' image classification datasets such as MNIST \citep{lecun-mnisthandwrittendigit-2010}, Cifar10, and Cifar100 \citep{krizhevsky2009learning}. Furthermore, we also include the Synbols \citep{NEURIPS2020_0169cf88} dataset, a character dataset composed of two independent labels: the characters and the fonts. (see \autoref{app:split_synbols} for the motivation).
Exhaustive results can be found at \url{https://wandb.ai/sequoia/csl_study}.

A sample of these results is illustrated in \autoref{fig:incremental_sl}, which shows results of various methods in the class-IL and task-IL settings in terms of their final performance and runtime. We note that some \codeword{Avalanche} methods achieve lower than chance accuracy in task-IL because they do not use the task label to mask out the classes that lie outside the tested task.

% \begin{minipage}{\textwidth}
%   \begin{minipage}{0.65\textwidth}
\begin{figure}
    \centering
    \includegraphics[width=0.335\linewidth]{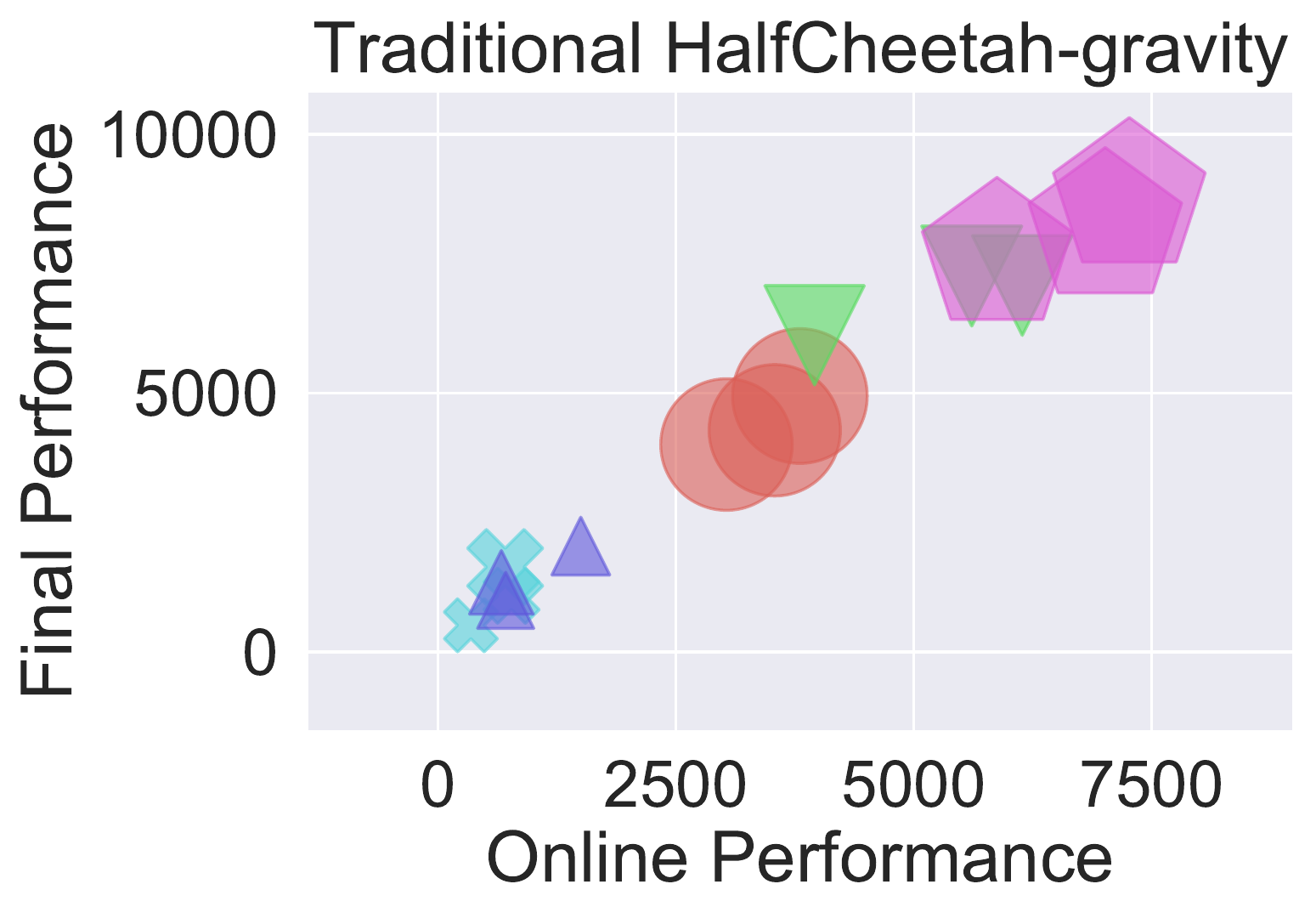} 
    \includegraphics[width=0.48\linewidth]{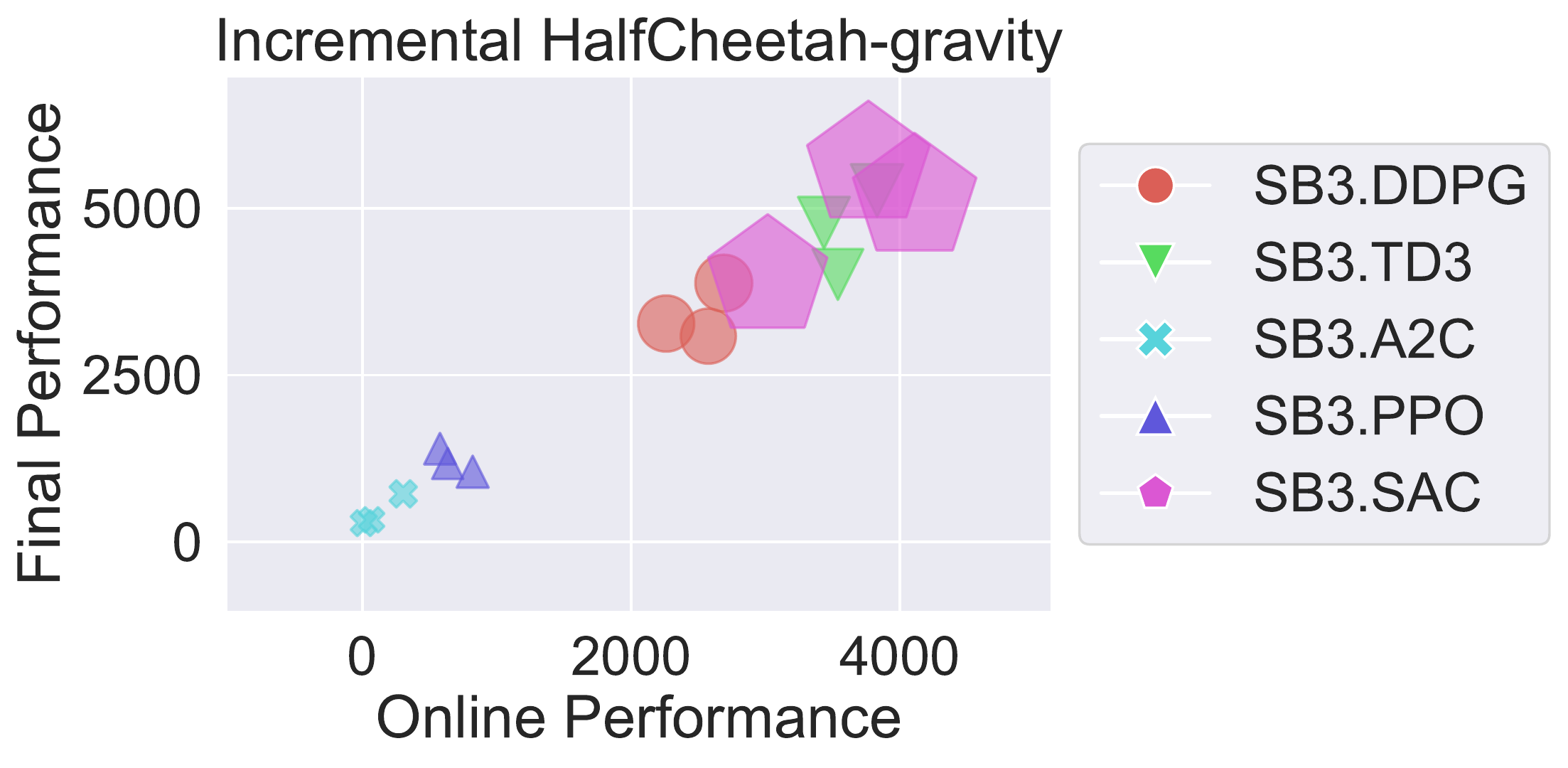} \\
    \includegraphics[width=0.335\linewidth]{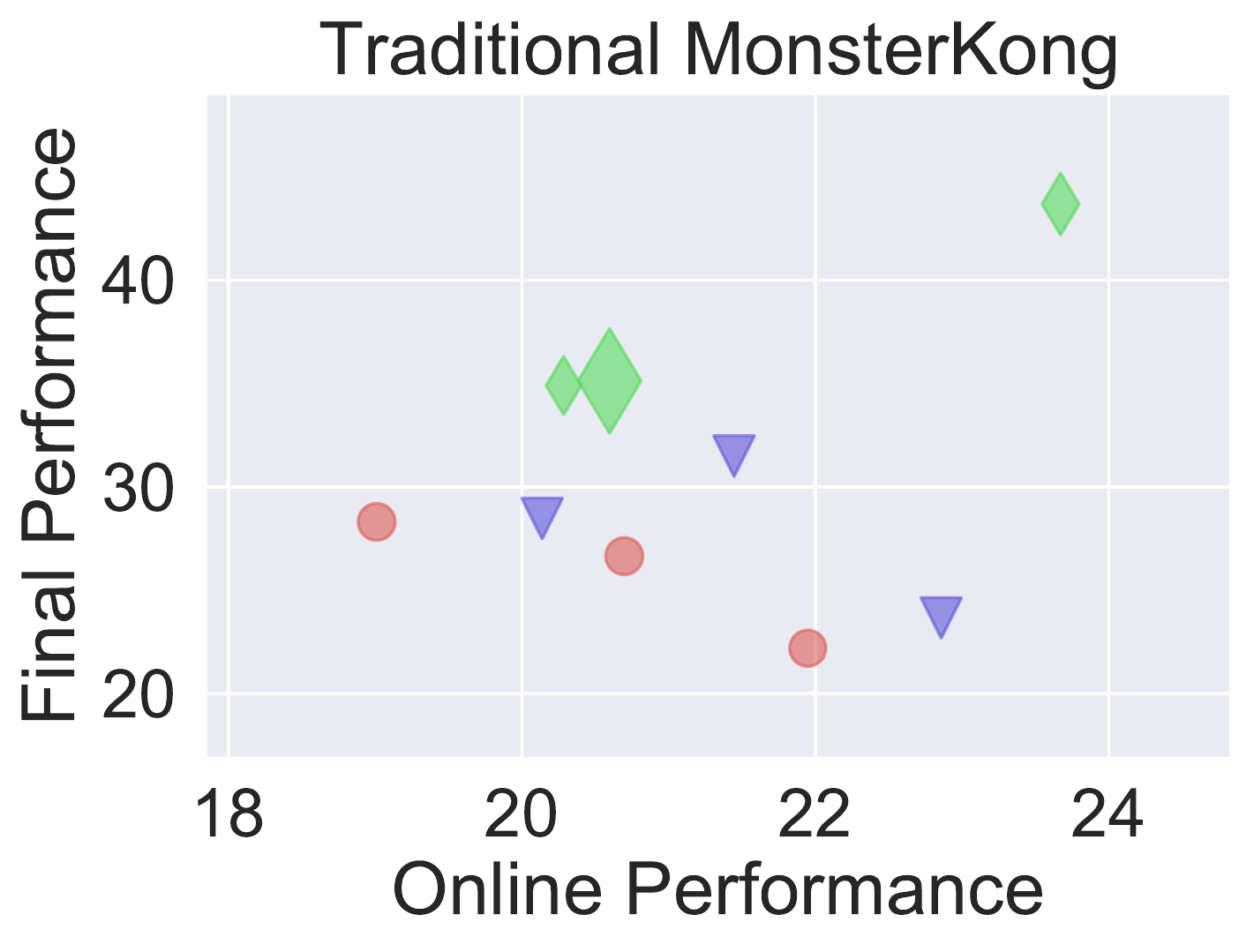} 
    \includegraphics[width=0.48\linewidth]{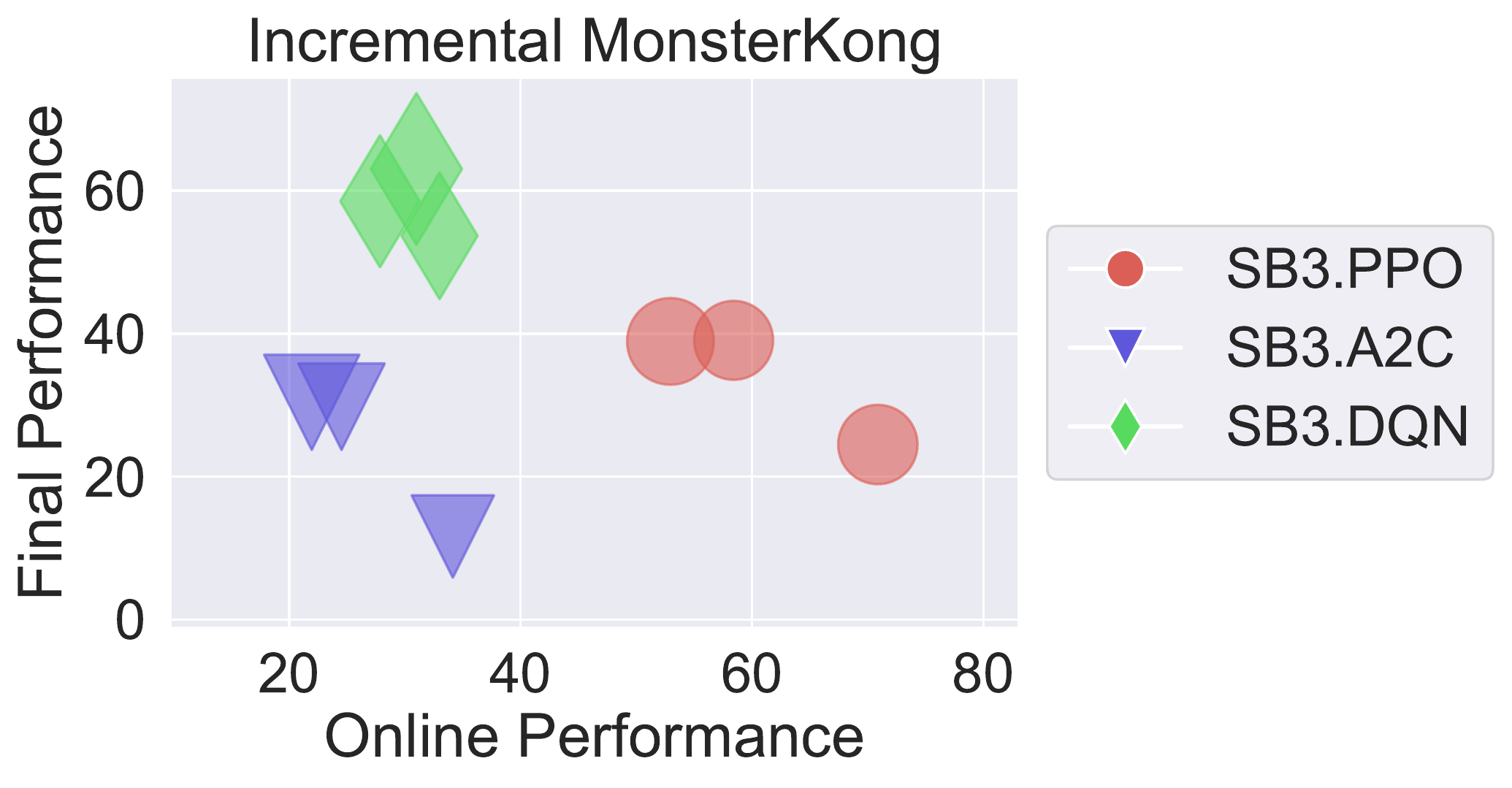} 
    \captionof{figure}{\textbf{Impact of the RL backbone algorithm in Traditional and Incremental RL}. Final performance (vertical axis) is plotted against online performance (horizontal axis). The bubbles' size indicates the normalized runtime of the methods. The methods achieving the best trade-off lie closer to the top-right of the figures and have smaller bubble size. Datasets are presented in each row and settings are presented in each column. For each method, several trials are presented depending on metrics composed of linear combination of final performance and online performance.} %Overall, SAC and TD3 seem like the superior methods in the continuous domain (HalfCheetah) and analogously PPO and A2C in the discrete domain (MonsterKong).}
    \label{fig:backbone_rl}
%   \end{minipage}
\end{figure}

%--------------------------------
\textbf{Continual Reinforcement Learning.}
We apply the RL methods from SB3 on multiple benchmarks built on HalfCheetah-v2, Hopper-v2, MountainCar-v0, CartPole-v0, MetaWorld-v2 (details in \autoref{app:benchmarks}). We also introduce a new discrete domain benchmark, namely Continual-MonsterKong, that we developed to study forward transfer in a more meaningful way (see  \autoref{app:continual_monsterkong} for more details). Complete results are available at \url{https://wandb.ai/sequoia/crl_study}.

A sample of these results is in \autoref{fig:backbone_rl}. It presents various methods in the traditional and incremental learning settings with their final performance, online performance and normalized runtime.

In \autoref{tbl:incremental_rl}, we apply the \codeword{continual-world} methods, built on top of SAC, on a incremental RL benchmark inspired by \citep{mendez2020lifelong} (see \autoref{app:mujoco-openai} for more details).
Finally, \autoref{fig:crlbenchmark_eval_rl} shows the transfer matrix achieved by one such algorithm, namely PPO~\citep{schulman2017proximal}.

%   \hspace{1.5ex}
%   \begin{minipage}[t!]{0.36\textwidth}
\begin{table}
    \centering
    % here: https://docs.google.com/spreadsheets/d/1rN5x1VkqJQmhc_ZaWX6vZxNHGnrD58zTIZOhiBV9Ai0/edit#gid=0
      \begin{small}
\begin{tabular}{llll}
\toprule
Method  & Final Perf.          & Online Perf.        & Runtime (h) \\
\midrule
SAC (base)    &   194 ± 105 &   254 ± 9 &  19.0±0.3 \\
AGEM    &   787 ± 268 &  283 ± 14 &  24.9±2.6 \\
EWC     &   616 ± 257 &  232 ± 17 &  19.3±2.5 \\
L2      &   840 ± 224 &  245 ± 22 &  18.7±2.5 \\
MAS     &   607 ± 236 &  236 ± 13 &  20.3±1.7 \\
PackNet &  1153 ± 325 &  \textbf{290 ± 41} &  25.2±6.3 \\
Perfect Memory    &  \textbf{1500 ± 399} &   255 ± 7 &  \textbf{18.2±2.0} \\
\bottomrule
\end{tabular}
    \end{small}
    \captionof{table}{\textbf{Incremental RL results.} Multiple CRL methods, all built on top of SAC, are tested on the Hopper-Bodyparts benchmarks. All CRL methods outperformed the Fine-tuning SAC baseline, validating their efficacy. Experience replay with a Perfect Memory achieves the best retained performance on all tasks, followed closely by PackNet.}
    \label{tbl:incremental_rl}
    % \end{flushright}
    % \end{minipage}
% \end{minipage}
\end{table}

\clearpage
%---------------------------------------------------------------------------------------------------
\section{Conclusion}

In this work, we introduce Sequoia: a publicly available framework to organize virtually all research settings from both the fields of Continual Supervised and Continual Reinforcement learning. Sequoia also makes methods are directly reusable by contract across settings using inheritance. It is our hope that Sequoia will be useful to new and experienced researchers in CL.
Further, the principles used to construct this framework for CL could very well be applied to other fields of research, effectively growing the tree towards new and interesting directions.
We welcome suggestions and contributions to that effect in our GitHub page at \href{https://www.github.com/lebrice/Sequoia}{github.com/lebrice/Sequoia}.

\section*{Acknowledgements}

We thank Samsung Electronics Co., Ldt. for their support.

\bibliographystyle{colla2022/collas2022_conference}
\bibliography{references}

\appendix

\input{appendix}

\end{document}

%% file: math_commands.tex
\usepackage{amsmath,amsfonts,bm}

\def\eqref#1{equation~\ref{#1}}
\def\1{\bm{1}}

\DeclareMathAlphabet{\mathsfit}{\encodingdefault}{\sfdefault}{m}{sl}
\SetMathAlphabet{\mathsfit}{bold}{\encodingdefault}{\sfdefault}{bx}{n}

%% file: appendix.tex
\appendix

\section{Reproducibility statement}

To facilitate reproducing results of our experiments, we include an anonymized version of the Sequoia codebase.
All results from the experiments of \autoref{sec:experiments} can be observed at \url{https://wandb.ai/sequoia}.
Each run includes the exact command used, as well as the git state, the complete system specification, hyper-parameter configurations, and random seeds used.

While most sources of randomness are accounted for in Sequoia, we are still in the process of making settings entirely deterministic given a random seed. In other words, for some combinations of settings and methods, launching two runs with the exact same arguments and seeds do sometimes produce different results. Making settings and methods entirely deterministic is part of the plans for future work in this project.

\newpage
\section{Supported methods}

One of Sequoia's biggest strength is how easy it is to extend. Most methods in Sequoia are the result directly reusing existing implementations from other frameworks and repositories, such as Avalanche\cite{lomonaco2021avalanche}, Stable-Baselines3\cite{stable-baselines3} and Continual World\cite{ContinualWorld}. \autoref{tab:full_support} shows all the methods currently available in Sequoia.

% External repositories can register their own methods through a simple plugin system.

\todo[inline]{in table 1, make alternating grey and with rows}
\begin{table}[h]
    \centering
    \begin{tabular}{l|l}
    % \begin{tabular}{l||p{\TableColWidth}p{\TableColWidth}p{\TableColWidth}p{\TableColWidth}p{\TableColWidth}p{\TableColWidth} | p{\TableColWidth}p{\TableColWidth}}
    % & \multicolumn{6}{p}{Continual learning assumptions} & \multicolumn{2}{|p}{SL vs RL} \\
    Method &  Target setting \\
    \hline
        \codeword{BaseMethod}                                       & \codeword{Setting} (all) \\
        \codeword{BaseMethod}.EWC \cite{kirkpatrick2017overcoming}  & \codeword{Setting} (all) \\
        \codeword{BaseMethod}.PackNet \cite{mallya2018packnet}      & Incremental Learning (RL + SL) \\
        % La-MAML \cite{gupta2020maml}                                & \\
        replay                                                          & Incremental SL \\
        CN-DPM \cite{CN-DPM}                                            & Continual SL \\
        HAT \cite{HAT_Serra}                                            &  Task-Incremental SL\\
        PNN \cite{Rusu16progressive}                                    &  Incremental SL \\
        \codeword{Avalanche}.naive \cite{lomonaco2021avalanche}         &  Incremental SL \\
        \codeword{Avalanche}.AGEM \cite{Chaudhry19}                     &  Incremental SL \\
        \codeword{Avalanche}.cwr\_star \cite{lomonaco2021avalanche}     &  Incremental SL \\
        \codeword{Avalanche}.EWC \cite{kirkpatrick2017overcoming}       &  Incremental SL \\
        \codeword{Avalanche}.Gdumb \cite{prabhu12356gdumb}              &  Incremental SL \\
        \codeword{Avalanche}.GEM \cite{lopez2017gradient}               &  Incremental SL \\
        \codeword{Avalanche}.LWF \cite{li2019continual}                 &  Incremental SL \\
        \codeword{Avalanche}.replay                                     &  Incremental SL \\
        \codeword{Avalanche}.SI \cite{Zenke17}                          &  Incremental SL \\
        \codeword{stable-baselines3}.A2C \cite{mnih2016asynchronous}    &  Incremental RL \\
        \codeword{stable-baselines3}.DDPG \cite{lillicrap2015continuous}    & Continual RL \\
        \codeword{stable-baselines3}.DQN \cite{mnih2015human}               & Continual RL \\
        \codeword{stable-baselines3}.PPO \cite{schulman2017proximal}        & Continual RL \\
        \codeword{stable-baselines3}.SAC \cite{haarnoja2018soft}            & Continual RL \\
        \codeword{stable-baselines3}.TD3 \cite{fujimoto2018addressing}      & Continual RL \\
        \codeword{continual_world}.SAC \cite{haarnoja2018soft}              & Incremental RL \\
        \codeword{continual_world}.AGEM \cite{Chaudhry19}                   & Incremental RL \\
        \codeword{continual_world}.EWC \cite{kirkpatrick2017overcoming}     & Incremental RL \\
        \codeword{continual_world}.VCL \cite{VCL}                           & Incremental RL \\
        \codeword{continual_world}.MAS \cite{MAS}                           & Incremental RL \\
        \codeword{continual_world}.L2 regularization                        & Incremental RL \\
        \codeword{continual_world}.PackNet \cite{mallya2018packnet}         & Incremental RL \\
        \codeword{continual_world}.Replay                                   & Incremental RL \\
         \vspace{0.2cm} \\
        \end{tabular}
    \caption{\label{tab:full_support}\textbf{Sequoia's methods support.} Each method specifies a target setting, listed on the right. Most methods are applicable in either RL or SL, while some can be applied to both. Methods also specify the ``level of nonstationarity" they are prepared to handle, as a choice of one of Continual (which is also referred to as Continuous Task-Agnostic CL in \autoref{sec:cl_assumption}), Discrete, Incremental, Task-Incremental, Traditional, and Multi-Task.} 
\end{table}

\section{The Base Method}
\label{app:base_method}

While developing a new \codeword{Method} in Sequoia, users are encouraged to separate the training logic from the networks used, the former being contained in the \codeword{Method}, and the latter in a model class, as advocated by PyTorch-Lightning~\cite{falcon2019pytorch} (PL), a powerful research library, which we employ as part of this \codeword{BaseMethod}.

The \codeword{BaseMethod} is accompanied by the \codeword{BaseModel}, which acts as a modular and extendable model for CL Methods to use. This \codeword{BaseModel} adheres to PyTorch-Lightning's \codeword{LightningModule} interface, making it easy to extend and customize with additional callbacks and loggers.
Likewise, the \codeword{BaseMethod} employs a \codeword{pl.Trainer}, which is able to train the \codeword{BaseModel} on the \codeword{Environment}s produced by any setting. Sequoia's Settings are also closely related to PL's \codeword{DataModule} abstraction. See \autoref{app:adding_new_settings} for a further discussion of the relationship between Sequoia and Pytorch-Lightning.

Using this \codeword{BaseModel} when creating a new CL method can be particularly useful when transitioning from a CL \codeword{Setting} to its parent, as it comes equipped with most of the components required to handle such transitions (e.g. task inference, multi-head prediction, etc.) These components, as well as the underlying encoder, output head, loss function, etc. can easily be replaced or customized.

Additional losses can also be added to the \codeword{BaseModel} through a modular interface, which was explicitly designed to facilitate exploration of self-supervised learning research.
% Examples of such auxiliary tasks include the EWC method, which is implemented as an extension of the \codeword{BaseMethod} which adds an EWC auxiliary task onto the \codeword{BaseModel}.

\section{Adding new Settings}
\label{app:adding_new_settings}

\begin{figure}
    \centering
    \includegraphics[width=\textwidth]{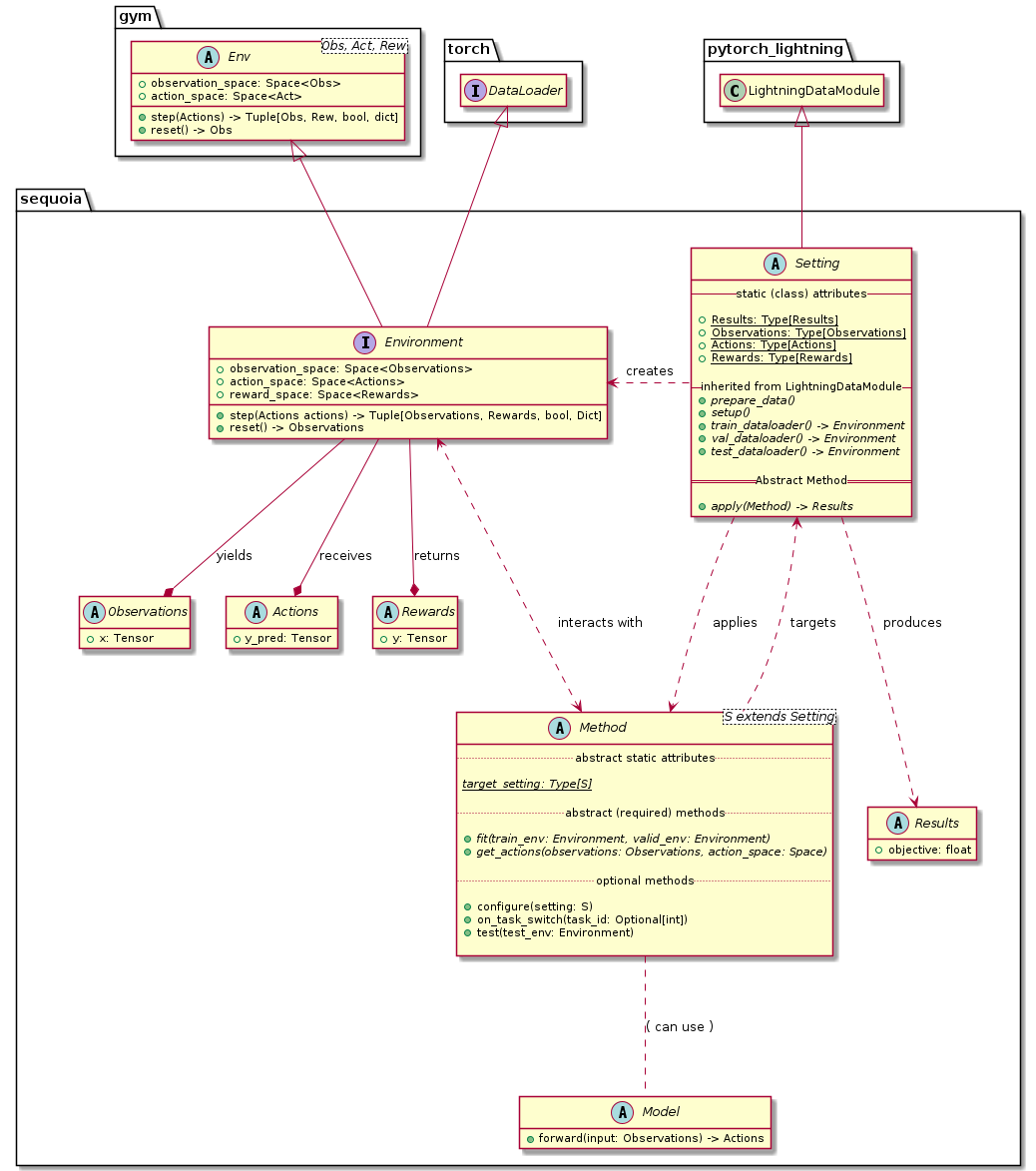}
    \caption{UML Diagram showing the main abstractions of Sequoia.}
    \label{fig:uml_base}
\end{figure}

There are three ways to easily create a new setting:

\begin{enumerate}
    \item By extracting an assumption present in the root setting, therefore creating a new \textit{root} or most general setting.

    For example, the most general CL setting we consider in this work, Continuous, Task-Agnostic Continual Learning (CTaCL) makes an implicit assumption that the non-stationarity of the environment isn’t affected by the actions of the agent.
    In this example, it could be argued that this kind of active non-stationarity is a more general form of non-stationarity than the passive variant. Therefore, it would follow that a method able to handle problems with active non-stationarity should also be applicable to problems with passive non-stationarity.
    This assumption could be extracted, to create a new (yet unamed) CL setting, with CTaCL as its child. Any Method that was previously declared to work in CTaCL will not be affected, and new methods aimed at this very challenging and general setting could be created to handle both types of settings.
    This is one of the major benefits of using an inheritance hierarchy to organize settings.

    \item Creating new leaves in the tree, by adding an assumption or constraints to an existing setting. One example of this could be settings where more information is available in the observations/actions/rewards than their parents.

    \item Adding new intermediate nodes to the tree, by making the differences in assumptions between existing settings more fine-grained: for example, if a jump between two settings is too large, a new intermediary node can be introduced, also without impacting the methods that were created for the parent or the child setting. 
 
\end{enumerate}
Finally, there is another, albeit more involved way to create new settings: to introduce a new class of assumptions. This separate assumption hierarchy is then composed with existing settings, resulting in a large increase in the number of settings.
For instance, if you consider the level of supervision, as-in, the availability of the rewards signal from the environment, you could recover unsupervised, semi-supervised, and “supervised”/traditional RL/SL settings. Through multiple inheritance, one could then create a new Setting for each combination of assumptions.

The need for each of these variations to be defined as classes is one shortcoming of the Sequoia framework as it currently stands, and will be addressed in future work using structural, rather than nominal subtyping.

\section{Relation between Sequoia and PyTorch-Lightning}
\label{app:sequoia_pl}

The very simple definition of the \codeword{Setting} class means that new Settings are not required to place themselves into our existing inheritance hierarchy, and could also not be related to continual learning at all!
It is however preferable, whenever possible, to find the closest existing Setting within Sequoia, and create the new Setting either below its closest relative or above it when adding or removing assumptions, respectively.

The most general Setting in our current hierarchy - also referred to as the "root`` setting - inherits from this abstract base class, while also building upon the elegant \codeword{DataModule} abstraction introduced by Pytorch-Lightning~\cite{falcon2019pytorch}, in which the \codeword{DataModule} is the entity responsible for the preparation of the data, as well as for the creation of the training, validation and testing \codeword{DataLoaders}. Models in pytorch lightning can thus easily train and evaluate themselves through this standardized API, where dataloaders can be swapped out between experiments.
Sequoia's main contributions can thus be viewed as taking this idea one step further, by 1) giving control of the ``main loop'' to this construct (through the addition of the \codeword{apply} method), 2) expanding this idea into the realm of Reinforcement Learning by moving from PyTorch's \codeword{DataLoaders} to a higher-level abstraction (\codeword{Environment}s), and 3) organizing these modules into an inheritance hierarchy.

Given how all current Sequoia Settings are instances of PL's \codeword{LightningDataModule} class, it is easy to use pytorch lightning for the training of Methods in Sequoia. This is one of the reasons why, for instance, the \codeword{BaseMethod} uses Pytorch Lightning's Trainer class in its implementation.
However, the trainer-based API is not directly usable, due to the very nature of CL problems, in which there are training dataloaders for each task, which isn't currently possible through the standard Pytorch Lightning's API.

\newpage
\section{Adding new Environments}
\label{app:add_new_env}
\providecommand*{\listingautorefname}{Listing}
New environments can be added to Sequoia by registering a new handler for creating new tasks, as can be seen in \autoref{fig:add_new_env} for continuous tasks, and in \autoref{fig:add_new_discrete_env} for discrete tasks.

\begin{listing*}
\scriptsize
\begin{minted}{python}
from typing import List, Dict
from sequoia.settings.rl import make_continuous_task
from gym.envs.classic_control import CartPoleEnv
import numpy as np

# A Continuous task is a dict mapping from attributes to the values
# to be set on the environment:
ContinuousTask = Dict[str, float]

@make_continuous_task.register(CartPoleEnv)
def make_task_for_my_env(
    env: CartPoleEnv, step: int, change_steps: List[int], seed: int = None, **kwargs,
) -> ContinuousTask:
    # NOTE: task sampling should be reproducible given a `seed`.
    step_seed = seed * step if seed is not None else None
    rng = np.random.default_rng(step_seed)
    return {
        "gravity": 9.8 * rng.normal(1, 0.5),
        "masscart": 1.0 * rng.normal(1, 0.5),
        "masspole": 0.1 * rng.normal(1, 0.5),
        "length": 0.5 * rng.normal(1, 0.5),
        "force_mag": 10.0 * rng.normal(1, 0.5),
        "tau": 0.02 * rng.normal(1, 0.5),
    }
\end{minted}
% Temporary solution:
% \lstinputlisting[frame=single,breakindent=.5\textwidth,frame=single,breaklines=true,style=mypython]{listings/adding_new_continuous_tasks.py}
\captionof{listing}{Example of how to add new RL environments to Sequoia. In this example, we register a function which will be used to sample continuous tasks for this environment, allowing it to become used as part of the Continuous Task-Agnostic Continual RL Setting and all of its descendants. }\label{fig:add_new_env}
\end{listing*}

\begin{listing*}
\scriptsize
\begin{minted}{python}
import operator
from typing import List, Dict, Union, Callable
import numpy as np
import gym
from metaworld.envs.mujoco.sawyer_xyz.v2 import SawyerReachEnvV2
from metaworld import ML10
from sequoia.settings.rl.discrete import make_discrete_task

# In the case of Discrete RL settings, you can either return a 
# 'continuous' task as before or a callable which will be 
# applied onto the environment when a task boundary
# is reached:
ContinuousTask = Dict[str, float]
IncrementalTask = Union[ContinuousTask, Callable[[gym.Env], None]]

@make_discrete_task.register(SawyerReachEnvV2)
def make_discrete_task_for_metaworld_env(
    env: SawyerReachEnvV2,
    step: int,
    change_steps: List[int],
    seed: int = None,
    **kwargs,
) -> IncrementalTask:
    benchmark = ML10(seed=seed)
    rng = np.random.default_rng(seed)
    some_metaworld_task = rng.choice(benchmark.train_tasks)
    # NOTE: Equivalent to the following, but has the benefit of 
    # being pickleable for vectorized envs:
    # return lambda env: env.set_task(some_metaworld_task)
    return operator.methodcaller("set_task", some_metaworld_task)
\end{minted}
% \lstinputlisting[frame=single,breakindent=.5\textwidth,frame=single,breaklines=true,style=mypython]{listings/adding_new_discrete_tasks.py}
\captionof{listing}{Example of how to add support for new RL environments to Sequoia in the case of discrete tasks, which are applied when a task boundary is reached. In this example, we register a function which will be used to sample discrete tasks for this environment, allowing it to become used as part of the Discrete Task-Agnostic RL setting and its descendants. }\label{fig:add_new_discrete_env}
\end{listing*}

\label{app:settings_uml}
\begin{figure*}
    \centering
    \includegraphics[width=\textwidth, height=0.95\textheight]{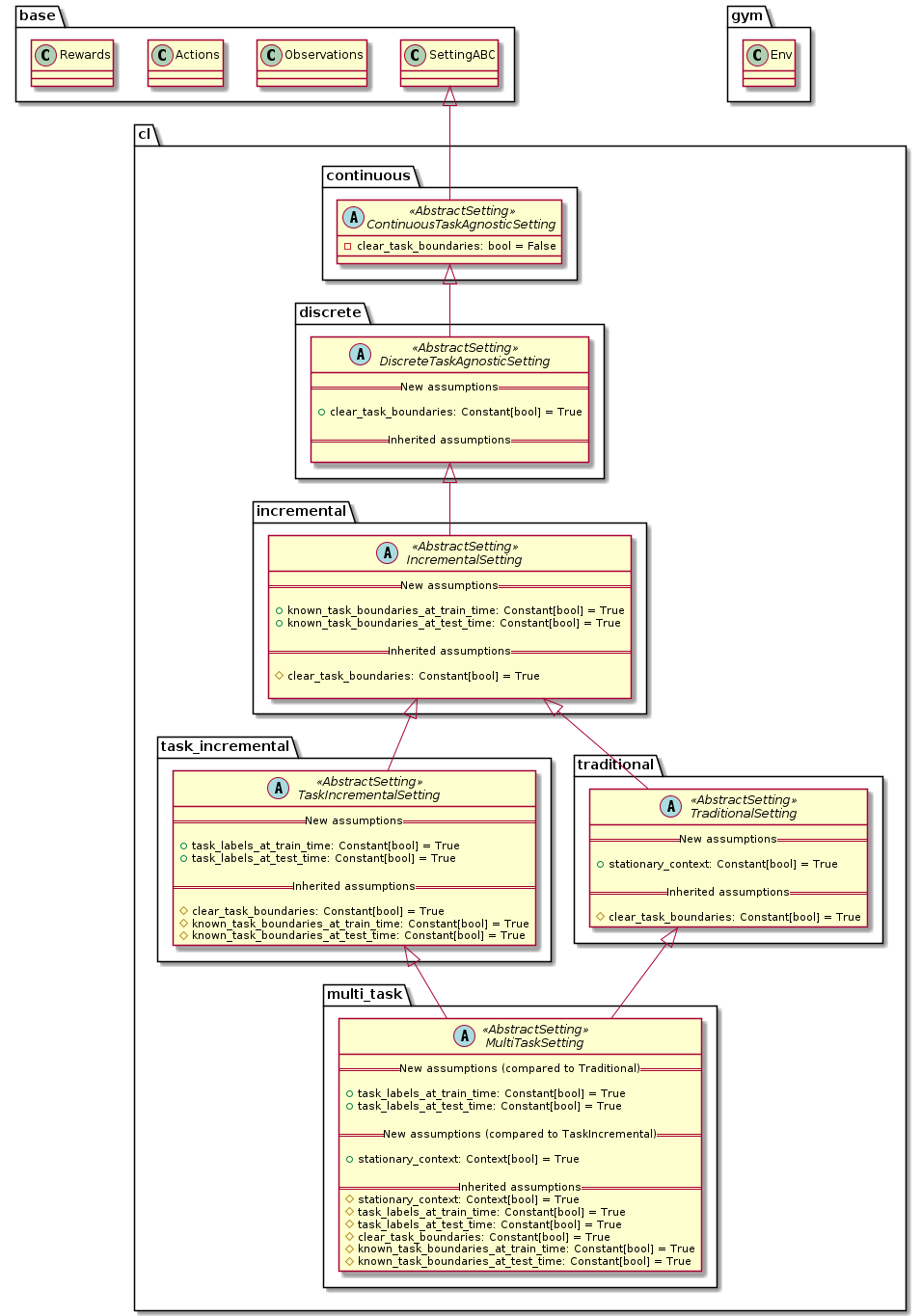}
    \caption{UML Diagram of the CL assumptions hierarchy. The CRL and CSL branches are not shown, but follow an identical structure.}
    \label{fig:uml_settings}
\end{figure*}

\clearpage
% ---------------------------------------
\section{Benchmark details}
\label{app:benchmarks}

% ---------------
\subsection{Split-Synbols dataset}
\label{app:split_synbols}

Currently employed datasets can't be used sensibly to construct domain-incremental learning problems. Some have used MNIST to construct Permuted-MNIST and Rotated-MNIST, however, \cite{Farquhar18} have explained and demonstrated why such benchmarks are flawed and bias their results unfairly towards some methods. Motivated by this, we introduce Split-Synbols. Based on the Synbols dataset~\cite{NEURIPS2020_0169cf88}, a character classification dataset in which examples have an extra label corresponding to their font, one can easily construct sensible domain-incremental benchmark where e.g., a font would consist of a domain. 

\begin{figure}[h]
  \centering
%   \vspace{-3mm}
    \includegraphics[width=1.0\linewidth]{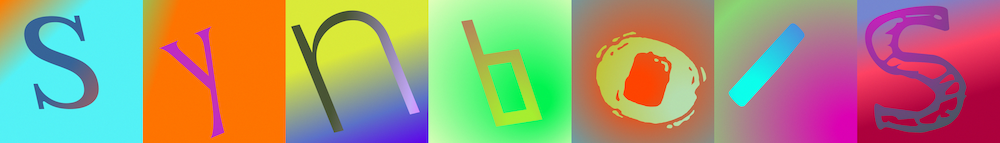}
    \caption{\textbf{Split-Synbols.} Example of Synbols character to classify. }
    \label{fig:split-synbol}
    % \vspace{-3mm}
\end{figure}

For the experiments, however, we opted for a class-incremental version to increase the difficulty. We prescribe a segmentation into 12 tasks to be learned sequentially, each consisting of a 4-way classification problem. Some example of Synbols character are displayed in Figure \ref{fig:split-synbol}.

%---------------
\subsection{Continual-MonsterKong environment}
\label{app:continual_monsterkong}

With rapid advancements in the field of deep RL, continual RL or \textit{never-ending}-RL has witnessed rekindled interest towards the goal for broad-AI in recent years.
While significant progress has been made in related domains such as transfer learning~\cite{JMLR09-taylor}, multi-task learning~\cite{ammar2014online, parisotto16_actormimic,calandriello2014multitask,maurerBenefitMultitaskRepresentation, barreto2019transfer, landolfi2019mbmtrl}, and generalization in RL~\cite{farebrother2018generalization}, an outstanding bottleneck is the lack of standard tools to develop and evaluate CRL agents~\cite{khetarpal2020towards}.
A standardized benchmark will potentially enable rapid research and development of CRL agents. % intro to CRL benchmark, why MonsterKong? 
To this end, we propose a new CRL benchmark within the unified framework of Sequoia. In particular, we build the CRL benchmark leveraging the Pygame learning environment \texttt{MonsterKong}~\cite{tasfi2016PLE}. MonsterKong is pixel-based, lightweight and has an easily-customizable domain, making it a good choice for evaluating continual learning agents. 

\begin{figure*}[t!]
  \centering
    \includegraphics[width=1.0\linewidth]{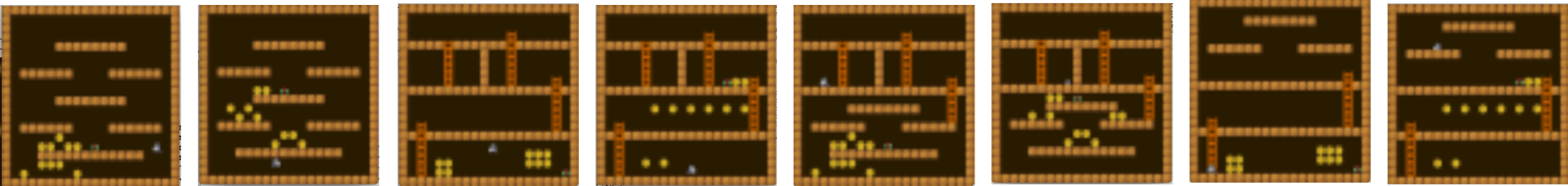}
    \caption{\textbf{Continual-MonsterKong.} We display the 8 tasks that constitute the benchmark in chronological order. The first two tasks test the agent's ability to jump between platforms, the second two test its ability to climb ladders and the last four combine both skills. }
    \label{fig:crlbenchmark}
\end{figure*}

Specifically, we design \emph{tasks} through a variety of map configurations. These configurations vary in terms of the location of the goal and the location of coins within each level. We introduce randomness across runs of a task by varying the start locations of the agent. To incorporate the ability to evaluate across specific CRL characteristics, we leverage tasks to define CRL \emph{experiments}. We design families of tasks leveraging the following abstract concepts: \textit{jumping tasks} which require the agent to perform jumps across platforms of different lengths in order to collect coins and reach the goal, \textit{climbing tasks} which require the agent to competently navigate ladders in order to collect coins and reach the goal, and tasks that combine both of these skills. The specific tasks leveraged as part of the CRL competition are depicted in Figure \ref{fig:crlbenchmark}. The agent trains on each task for 200,000 steps. 

\textbf{Experiment Details:} To evaluate the agents on the CRL benchmark, we follow the standard evaluation introduced above. Final performance reports accumulated reward per episode on all test environments, averaged over all tasks, after the end of training, whereas online performance is measured as the accumulated reward per episode on the training environment of the current task during training of all tasks. For the runtime score, we use set $\textit{max\_runtime}$ of 12 hours and $\textit{min\_runtime}$ to 1.5 hours. Lastly, the agents are allowed a maximum of 200,000 steps per task.

\textbf{Customization:} Ideally, CRL agents must be able to solve tasks by acquiring knowledge in the form of skills, be able to use previously acquired behaviors, and build even more complex behaviours over the course of its lifetime~\cite{ring1997child, thrun1995lifelong, thrun1995finding}. While leveraging the MonsterKong environment, it is easy to introduce new environment layouts or modifications to existing layouts. Configurations could be customized to include arbitrary configurations of coins, ladders, platforms, walls, monsters, fire balls, and spikes. Making custom environment elements is straightforward as well, so the environment can be modified to aligned with the properties of the CRL agent that we would like to test. 

While in our benchmark we mainly focused on three families of tasks within the Monsterkong domain, it is fairly straightforward to introduce variations of map configurations to the framework. Monsterkong provides two degrees of design choices 1. the task definitions and 2. the evolution of tasks referred to as experiment definitions. Due to the nature of how tasks are specified through simple matrices (map configurations), many layers of complexity can be added through the task specification. For example, object addition and removal can induce local variations in reward, nails can be penalizing, diamonds can be bonuses. Additionally, changes to the textures of the game like simple changes to the color of the walls, the coins, and the background as well as changes in the lighting are easy to add for users interested to test generalization of the policies learned. 

%---------------
\subsection{HalfCheetah-gravity and Hopper-Bodyparts}
\label{app:mujoco-openai}

HalfCheetah-gravity and Hopper-Bodyparts are two benchmarks introduced in \cite{mendez2020lifelong}. In the first, each task consist of a different gravity. In the latter, the agent's body parts are changing in size at each tasks. The gravity and body parts values are sampled as in \cite{mendez2020lifelong}. The two benchmarks we study are each composed of 10 tasks.
\autoref{fig:app_backbone_rl} shows the results of this study.

\clearpage
% ---------------------------------------
\section{Extended Experiments}
\label{app:ext_exp}

\begin{figure}[h]
  \centering
%   \vspace{-3mm}
    \includegraphics[width=0.37\linewidth]{figures/sl/setting=Class-Incremental_dataset=MNIST.pdf} 
    \includegraphics[width=0.62\linewidth]{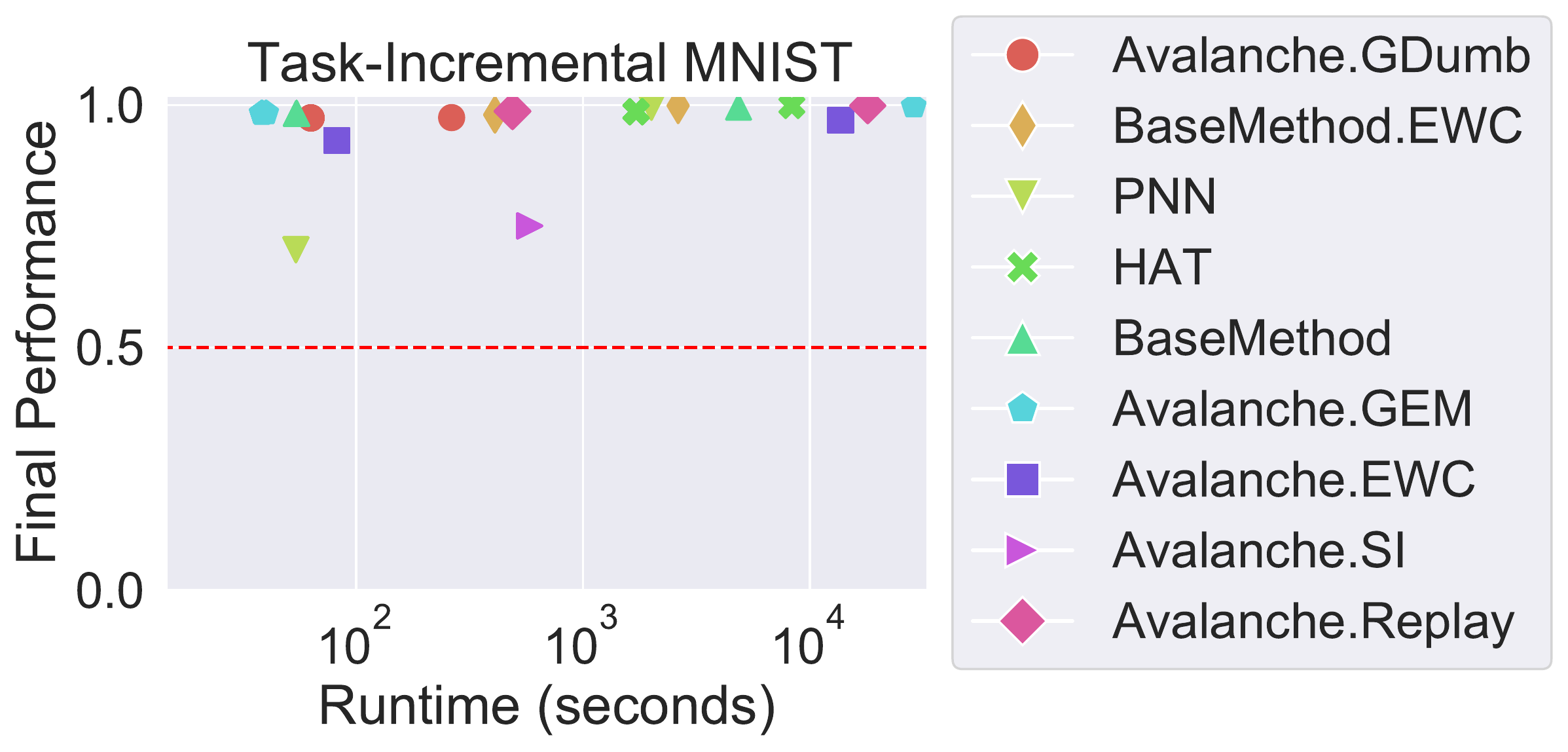} 
    \includegraphics[width=0.37\linewidth]{figures/sl/setting=Class-Incremental_dataset=CIFAR10.pdf} 
    \includegraphics[width=0.62\linewidth]{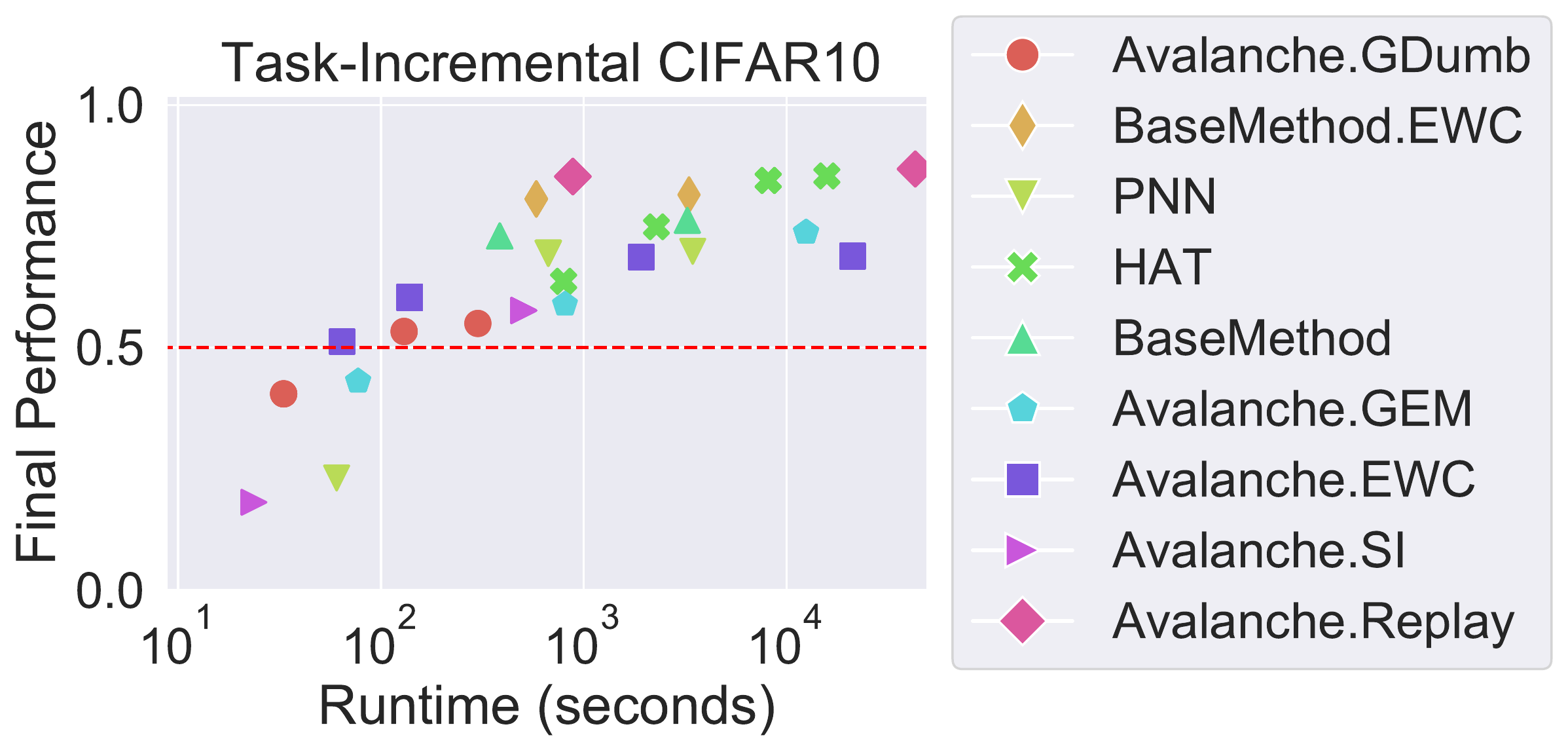} 
    \includegraphics[width=0.37\linewidth]{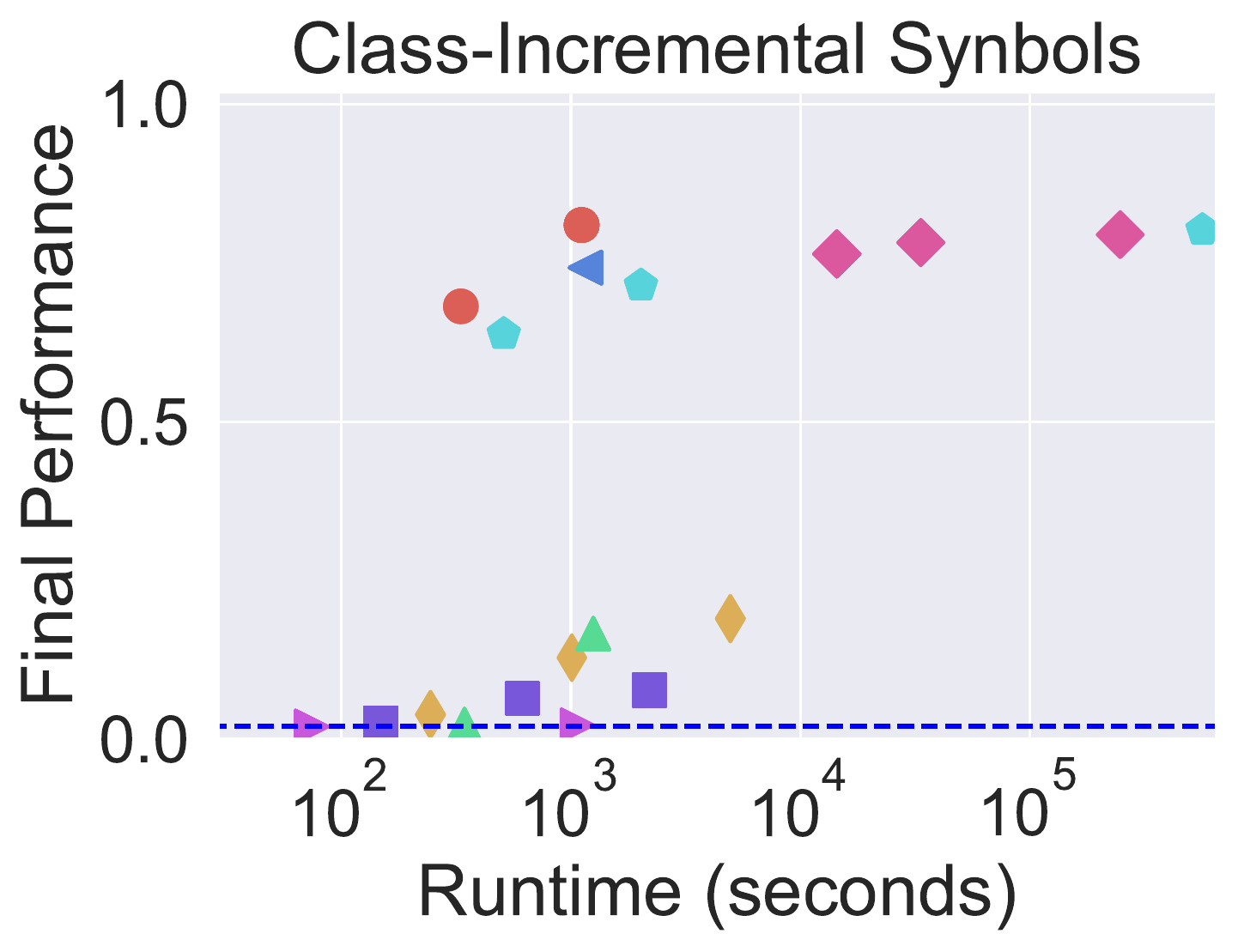} 
    \includegraphics[width=0.62\linewidth]{figures/sl/setting=Task-Incremental_dataset=Synbols_legend.pdf} 

    \caption{\textbf{Incremental Supervised Learning results}. Transpose of \autoref{fig:incremental_sl} for improved readability. Final performance (vertical axis) is plotted against runtime (horizontal axis). The methods achieving the best trade-off lie closer to the top-left of the figures.  The dotted line shows chance accuracy for each setting-dataset combination. For each methods, several trials are presented depending on metrics composed of linear combination of final performance and (normalized) runtime. Intuitively, better performance in CL normally comes at the cost of increased computation. This intuition is reflected in the presented results, as highlighted by the observed correlation between final performance and runtime. GEM and GDumb achieve the best tradoff, although the latter cannot make predictions in an online manner and thus serves more as a reference point. }
    \label{fig:app_incremental_sl}
    % \vspace{-3mm}
\end{figure}

% \todo{add some analysis}
\begin{figure}
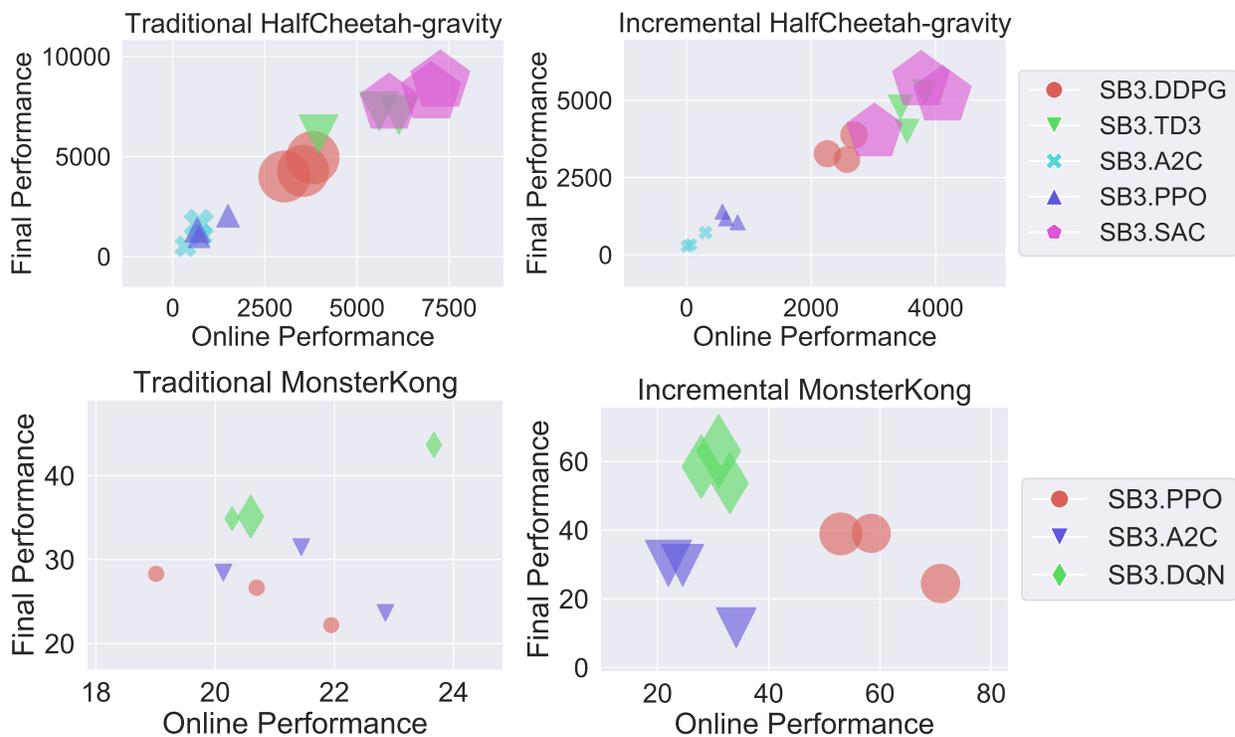

    % \centering
    \includegraphics[width=0.41\linewidth]{figures/rl/setting=Traditional_dataset=HalfCheetah-gravity} 
    \includegraphics[width=0.585\linewidth]{figures/rl/setting=Incremental_dataset=HalfCheetah-gravity_legend.pdf} \\
    \includegraphics[width=0.41\linewidth]{figures/rl/monsterkong/setting=Traditional_dataset=MonsterKong.pdf} 
    \includegraphics[width=0.585\linewidth]{figures/rl/monsterkong/setting=Incremental_dataset=MonsterKong_legend.pdf} 
    \captionof{figure}{\textbf{Impact of the RL backbone algorithm in Traditional and Incremental RL}. Larger version of \autoref{fig:backbone_rl} for improved readability. Final performance (vertical axis) is plotted against online performance (horizontal axis). The bubbles' size indicates the normalized runtime of the methods. Datasets are presented in each row and settings are presented in each column. For each method, several trials are presented depending on metrics composed of linear combination of final performance and online performance. The methods achieving the best trade-off lie closer to the top-right of the figures and have smaller bubble size. In general, we observe a trade-off between performance and runtime, a tendency also observed in \autoref{fig:app_incremental_sl}. Another interesting trade-off can be observed between final performance and online performance. E.g., in both MonsterKong benchmarks, DQN achieves the best final performance whereas PPO achieves the best online performance. Because the former is off-policy, it can re-use the previously acquired data to retain its performance on past tasks, increasing final performance. Contrarily, the latter, being on-policy, focuses on the current task and thus learns it faster, thereby increasing its online performance.}
    \label{fig:app_backbone_rl}
\end{figure}

% Tranfer Matrix
\begin{figure}[h!]
  \centering
%   \vspace{-3mm}
    \includegraphics[width=0.7\linewidth]{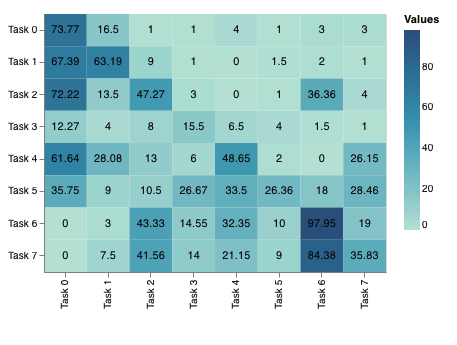} 
    \caption{\textbf{PPO's Transfer matrix in Continual-MonsterKong.} Each cell at row $i$ and column $j$ indicates the test performance on task $j$ after having learned tasks $0$ through $i$. The contents of each cell correspond to the average reward per episode obtained in the test environment for the corresponding task. Positive numbers above the diagonal indicate generalization to unseen tasks, which is achievable by design in the Continual-Monsterkong benchmark. }
    \label{fig:crlbenchmark_eval_rl}
    % \vspace{-3mm}
\end{figure}

%% file: main.bbl
\begin{thebibliography}{70}
\providecommand{\natexlab}[1]{#1}
\providecommand{\url}[1]{\texttt{#1}}
\expandafter\ifx\csname urlstyle\endcsname\relax
  \providecommand{\doi}[1]{doi: #1}\else
  \providecommand{\doi}{doi: \begingroup \urlstyle{rm}\Url}\fi

\bibitem[Aljundi et~al.(2018)Aljundi, Babiloni, Elhoseiny, Rohrbach, and
  Tuytelaars]{MAS}
Rahaf Aljundi, Francesca Babiloni, Mohamed Elhoseiny, Marcus Rohrbach, and
  Tinne Tuytelaars.
\newblock Memory aware synapses: Learning what (not) to forget, 2018.

\bibitem[Aljundi et~al.(2019{\natexlab{a}})Aljundi, Kelchtermans, and
  Tuytelaars]{Aljundi2019TaskFreeCL}
Rahaf Aljundi, Klaas Kelchtermans, and Tinne Tuytelaars.
\newblock Task-free continual learning.
\newblock \emph{2019 IEEE/CVF Conference on Computer Vision and Pattern
  Recognition (CVPR)}, pp.\  11246--11255, 2019{\natexlab{a}}.

\bibitem[Aljundi et~al.(2019{\natexlab{b}})Aljundi, Lin, Goujaud, and
  Bengio]{Aljundi2019Gradient}
Rahaf Aljundi, Min Lin, Baptiste Goujaud, and Yoshua Bengio.
\newblock Gradient based sample selection for online continual learning.
\newblock In \emph{Advances in Neural Information Processing Systems
  (NeurIPS)}, 2019{\natexlab{b}}.

\bibitem[Ammar et~al.(2014)Ammar, Eaton, Ruvolo, and Taylor]{ammar2014online}
Haitham~Bou Ammar, Eric Eaton, Paul Ruvolo, and Matthew Taylor.
\newblock Online multi-task learning for policy gradient methods.
\newblock In \emph{International conference on machine learning}, pp.\
  1206--1214, 2014.

\bibitem[Barreto et~al.(2019)Barreto, Borsa, Quan, Schaul, Silver, Hessel,
  Mankowitz, {\v{Z}}{\'\i}dek, and Munos]{barreto2019transfer}
Andr{\'e} Barreto, Diana Borsa, John Quan, Tom Schaul, David Silver, Matteo
  Hessel, Daniel Mankowitz, Augustin {\v{Z}}{\'\i}dek, and Remi Munos.
\newblock Transfer in deep reinforcement learning using successor features and
  generalised policy improvement.
\newblock \emph{arXiv preprint arXiv:1901.10964}, 2019.

\bibitem[Brockman et~al.(2016)Brockman, Cheung, Pettersson, Schneider,
  Schulman, Tang, and Zaremba]{brockman2016openai}
Greg Brockman, Vicki Cheung, Ludwig Pettersson, Jonas Schneider, John Schulman,
  Jie Tang, and Wojciech Zaremba.
\newblock Openai gym.
\newblock \emph{arXiv preprint arXiv:1606.01540}, 2016.

\bibitem[Caccia et~al.(2020)Caccia, Rodriguez, Ostapenko, Normandin, Lin,
  Page-Caccia, Laradji, Rish, Lacoste, V{\'a}zquez, et~al.]{caccia2020online}
Massimo Caccia, Pau Rodriguez, Oleksiy Ostapenko, Fabrice Normandin, Min Lin,
  Lucas Page-Caccia, Issam~Hadj Laradji, Irina Rish, Alexandre Lacoste, David
  V{\'a}zquez, et~al.
\newblock Online fast adaptation and knowledge accumulation (osaka): a new
  approach to continual learning.
\newblock \emph{Advances in Neural Information Processing Systems}, 33, 2020.

\bibitem[Calandriello et~al.(2014)Calandriello, Lazaric, and
  Restelli]{calandriello2014multitask}
Daniele Calandriello, Alessandro Lazaric, and Marcello Restelli.
\newblock Sparse multi-task reinforcement learning.
\newblock In Z.~Ghahramani, M.~Welling, C.~Cortes, N.~D. Lawrence, and K.~Q.
  Weinberger (eds.), \emph{Advances in neural information processing systems
  27}, pp.\  819--827. Curran Associates, Inc., 2014.

\bibitem[Chandak et~al.(2020{\natexlab{a}})Chandak, Theocharous, Nota, and
  Thomas]{chandak2020lifelong}
Yash Chandak, Georgios Theocharous, Chris Nota, and Philip~S. Thomas.
\newblock Lifelong learning with a changing action set, 2020{\natexlab{a}}.

\bibitem[Chandak et~al.(2020{\natexlab{b}})Chandak, Theocharous, Shankar,
  Mahadevan, White, and Thomas]{chandakfuture}
Yash Chandak, Georgios Theocharous, Shiv Shankar, Sridhar Mahadevan, Martha
  White, and Philip~S Thomas.
\newblock Optimizing for the future in non-stationary mdps.
\newblock \emph{arXiv preprint arXiv:2005.08158}, 2020{\natexlab{b}}.

\bibitem[Chaudhry et~al.(2019)Chaudhry, Ranzato, Rohrbach, and
  Elhoseiny]{Chaudhry19}
Arslan Chaudhry, Marc’Aurelio Ranzato, Marcus Rohrbach, and Mohamed
  Elhoseiny.
\newblock Efficient lifelong learning with {A-GEM}.
\newblock In \emph{International Conference of Learning Representations
  (ICLR)}, 2019.

\bibitem[Choi et~al.(2000)Choi, Yeung, and Zhang]{choi2000hidden}
Samuel~PM Choi, Dit-Yan Yeung, and Nevin~L Zhang.
\newblock Hidden-mode markov decision processes for nonstationary sequential
  decision making.
\newblock In \emph{Sequence Learning}, pp.\  264--287. Springer, 2000.

\bibitem[Delange et~al.(2021)Delange, Aljundi, Masana, Parisot, Jia, Leonardis,
  Slabaugh, and Tuytelaars]{delange2021continual}
Matthias Delange, Rahaf Aljundi, Marc Masana, Sarah Parisot, Xu~Jia, Ales
  Leonardis, Greg Slabaugh, and Tinne Tuytelaars.
\newblock A continual learning survey: Defying forgetting in classification
  tasks.
\newblock \emph{IEEE Transactions on Pattern Analysis and Machine
  Intelligence}, 2021.

\bibitem[Douillard \& Lesort(2021)Douillard and Lesort]{douillard2021continuum}
Arthur Douillard and Timothée Lesort.
\newblock Continuum: Simple management of complex continual learning scenarios,
  2021.

\bibitem[{Falcon et al.}(2019)]{falcon2019pytorch}
William {Falcon et al.}
\newblock Pytorch lightning.
\newblock \emph{GitHub. Note:
  https://github.com/PyTorchLightning/pytorch-lightning}, 3, 2019.

\bibitem[Farebrother et~al.(2018)Farebrother, Machado, and
  Bowling]{farebrother2018generalization}
Jesse Farebrother, Marlos~C Machado, and Michael Bowling.
\newblock Generalization and regularization in dqn.
\newblock \emph{arXiv preprint arXiv:1810.00123}, 2018.

\bibitem[Farquhar \& Gal(2018)Farquhar and Gal]{Farquhar18}
Sebastian Farquhar and Yarin Gal.
\newblock Towards robust evaluations of continual learning.
\newblock \emph{arXiv preprint arXiv:1805.09733}, 2018.

\bibitem[Fernando et~al.(2017)Fernando, Banarse, Blundell, Zwols, Ha, Rusu,
  Pritzel, and Wierstra]{fernando2017pathnet}
Chrisantha Fernando, Dylan Banarse, Charles Blundell, Yori Zwols, David Ha,
  Andrei~A Rusu, Alexander Pritzel, and Daan Wierstra.
\newblock Pathnet: Evolution channels gradient descent in super neural
  networks.
\newblock \emph{arXiv preprint arXiv:1701.08734}, 2017.

\bibitem[French(1999)]{French99}
Robert~M. French.
\newblock Catastrophic forgetting in connectionist networks.
\newblock \emph{Trends in Cognitive Sciences}, 3\penalty0 (4):\penalty0
  128--135, 1999.
\newblock ISSN 13646613.
\newblock \doi{10.1016/S1364-6613(99)01294-2}.
\newblock URL
  \url{https://www.sciencedirect.com/science/article/abs/pii/S1364661399012942}.

\bibitem[Fujimoto et~al.(2018)Fujimoto, Hoof, and
  Meger]{fujimoto2018addressing}
Scott Fujimoto, Herke Hoof, and David Meger.
\newblock Addressing function approximation error in actor-critic methods.
\newblock In \emph{International Conference on Machine Learning}, pp.\
  1587--1596. PMLR, 2018.

\bibitem[Haarnoja et~al.(2018)Haarnoja, Zhou, Abbeel, and
  Levine]{haarnoja2018soft}
Tuomas Haarnoja, Aurick Zhou, Pieter Abbeel, and Sergey Levine.
\newblock Soft actor-critic: Off-policy maximum entropy deep reinforcement
  learning with a stochastic actor.
\newblock In \emph{International Conference on Machine Learning}, pp.\
  1861--1870. PMLR, 2018.

\bibitem[Harrison et~al.(2019)Harrison, Sharma, Finn, and
  Pavone]{Harrison2019ContinuousMW}
James Harrison, Apoorva Sharma, Chelsea Finn, and Marco Pavone.
\newblock Continuous meta-learning without tasks.
\newblock \emph{ArXiv}, abs/1912.08866, 2019.

\bibitem[He et~al.(2019)He, Sygnowski, Galashov, Rusu, Teh, and
  Pascanu]{He2019TaskAC}
Xu~He, Jakub Sygnowski, Alexandre Galashov, Andrei~A. Rusu, Yee~Whye Teh, and
  Razvan Pascanu.
\newblock Task agnostic continual learning via meta learning.
\newblock \emph{ArXiv}, abs/1906.05201, 2019.

\bibitem[Henderson et~al.(2019)Henderson, Islam, Bachman, Pineau, Precup, and
  Meger]{henderson2019deep}
Peter Henderson, Riashat Islam, Philip Bachman, Joelle Pineau, Doina Precup,
  and David Meger.
\newblock Deep reinforcement learning that matters, 2019.

\bibitem[Kaplanis et~al.(2020)Kaplanis, Clopath, and
  Shanahan]{kaplanis2020continual}
Christos Kaplanis, Claudia Clopath, and Murray Shanahan.
\newblock Continual reinforcement learning with multi-timescale replay.
\newblock \emph{arXiv preprint arXiv:2004.07530}, 2020.

\bibitem[Khetarpal et~al.(2018)Khetarpal, Ahmed, Cianflone, Islam, and
  Pineau]{khetarpal2018re}
Khimya Khetarpal, Zafarali Ahmed, Andre Cianflone, Riashat Islam, and Joelle
  Pineau.
\newblock Re-evaluate: Reproducibility in evaluating reinforcement learning
  algorithms.
\newblock 2018.

\bibitem[Khetarpal et~al.(2020)Khetarpal, Riemer, Rish, and
  Precup]{khetarpal2020towards}
Khimya Khetarpal, Matthew Riemer, Irina Rish, and Doina Precup.
\newblock Towards continual reinforcement learning: A review and perspectives.
\newblock \emph{arXiv preprint arXiv:2012.13490}, 2020.

\bibitem[Kirkpatrick et~al.(2017)Kirkpatrick, Pascanu, Rabinowitz, Veness,
  Desjardins, Rusu, Milan, Quan, Ramalho, Grabska-Barwinska,
  et~al.]{kirkpatrick2017overcoming}
James Kirkpatrick, Razvan Pascanu, Neil Rabinowitz, Joel Veness, Guillaume
  Desjardins, Andrei~A Rusu, Kieran Milan, John Quan, Tiago Ramalho, Agnieszka
  Grabska-Barwinska, et~al.
\newblock Overcoming catastrophic forgetting in neural networks.
\newblock \emph{Proceedings of the national academy of sciences}, 114\penalty0
  (13):\penalty0 3521--3526, 2017.

\bibitem[Krizhevsky et~al.(2009)Krizhevsky, Hinton,
  et~al.]{krizhevsky2009learning}
Alex Krizhevsky, Geoffrey Hinton, et~al.
\newblock Learning multiple layers of features from tiny images.
\newblock Technical report, Citeseer, 2009.

\bibitem[Lacoste et~al.(2020)Lacoste, Rodr\'{\i}guez~L\'{o}pez,
  Branchaud-Charron, Atighehchian, Caccia, Laradji, Drouin, Craddock, Charlin,
  and V\'{a}zquez]{NEURIPS2020_0169cf88}
Alexandre Lacoste, Pau Rodr\'{\i}guez~L\'{o}pez, Frederic Branchaud-Charron,
  Parmida Atighehchian, Massimo Caccia, Issam~Hadj Laradji, Alexandre Drouin,
  Matthew Craddock, Laurent Charlin, and David V\'{a}zquez.
\newblock Synbols: Probing learning algorithms with synthetic datasets.
\newblock In H.~Larochelle, M.~Ranzato, R.~Hadsell, M.~F. Balcan, and H.~Lin
  (eds.), \emph{Advances in Neural Information Processing Systems}, volume~33,
  pp.\  134--146. Curran Associates, Inc., 2020.
\newblock URL
  \url{https://proceedings.neurips.cc/paper/2020/file/0169cf885f882efd795951253db5cdfb-Paper.pdf}.

\bibitem[Landolfi et~al.(2019)Landolfi, Thomas, and Ma]{landolfi2019mbmtrl}
Nicholas~C. Landolfi, Garrett Thomas, and Tengyu Ma.
\newblock A model-based approach for sample-efficient multi-task reinforcement
  learning, 2019.

\bibitem[LeCun \& Cortes(2010)LeCun and
  Cortes]{lecun-mnisthandwrittendigit-2010}
Yann LeCun and Corinna Cortes.
\newblock {MNIST} handwritten digit database.
\newblock http://yann.lecun.com/exdb/mnist/, 2010.

\bibitem[Lee et~al.(2020)Lee, Ha, Zhang, and Kim]{CN-DPM}
Soochan Lee, Junsoo Ha, Dongsu Zhang, and Gunhee Kim.
\newblock A neural dirichlet process mixture model for task-free continual
  learning, 2020.

\bibitem[Lesort et~al.(2019{\natexlab{a}})Lesort, Caselles-Dupr{\'e},
  Garcia-Ortiz, Goudou, and Filliat]{lesort2018generative}
Timoth{\'e}e Lesort, Hugo Caselles-Dupr{\'e}, Michael Garcia-Ortiz,
  Jean-Fran{\c c}ois Goudou, and David Filliat.
\newblock {Generative Models from the perspective of Continual Learning}.
\newblock In \emph{{International Joint Conference on Neural Networks
  (IJCNN)}}, 2019{\natexlab{a}}.

\bibitem[Lesort et~al.(2019{\natexlab{b}})Lesort, Gepperth, Stoian, and
  Filliat]{lesort2018marginal}
Timoth{\'e}e Lesort, Alexander Gepperth, Andrei Stoian, and David Filliat.
\newblock Marginal replay vs conditional replay for continual learning.
\newblock In \emph{International Conference on Artificial Neural Networks},
  pp.\  466--480. Springer, 2019{\natexlab{b}}.
\newblock URL \url{https://arxiv.org/abs/1810.12069}.

\bibitem[Lesort et~al.(2020)Lesort, Lomonaco, Stoian, Maltoni, Filliat, and
  Díaz-Rodríguez]{lesort2019continual}
Timothée Lesort, Vincenzo Lomonaco, Andrei Stoian, Davide Maltoni, David
  Filliat, and Natalia Díaz-Rodríguez.
\newblock Continual learning for robotics: Definition, framework, learning
  strategies, opportunities and challenges.
\newblock \emph{Information Fusion}, 58:\penalty0 52 -- 68, 2020.
\newblock ISSN 1566-2535.
\newblock \doi{https://doi.org/10.1016/j.inffus.2019.12.004}.
\newblock URL
  \url{http://www.sciencedirect.com/science/article/pii/S1566253519307377}.

\bibitem[Li et~al.(2019)Li, Barnaghi, Enshaeifar, and Ganz]{li2019continual}
HongLin Li, Payam Barnaghi, Shirin Enshaeifar, and Frieder Ganz.
\newblock Continual learning using bayesian neural networks, 2019.

\bibitem[Lillicrap et~al.(2015)Lillicrap, Hunt, Pritzel, Heess, Erez, Tassa,
  Silver, and Wierstra]{lillicrap2015continuous}
Timothy~P Lillicrap, Jonathan~J Hunt, Alexander Pritzel, Nicolas Heess, Tom
  Erez, Yuval Tassa, David Silver, and Daan Wierstra.
\newblock Continuous control with deep reinforcement learning.
\newblock \emph{arXiv preprint arXiv:1509.02971}, 2015.

\bibitem[Lomonaco et~al.(2021)Lomonaco, Pellegrini, Cossu, Carta, Graffieti,
  Hayes, Lange, Masana, Pomponi, van~de Ven, Mundt, She, Cooper, Forest,
  Belouadah, Calderara, Parisi, Cuzzolin, Tolias, Scardapane, Antiga, Amhad,
  Popescu, Kanan, van~de Weijer, Tuytelaars, Bacciu, and
  Maltoni]{lomonaco2021avalanche}
Vincenzo Lomonaco, Lorenzo Pellegrini, Andrea Cossu, Antonio Carta, Gabriele
  Graffieti, Tyler~L. Hayes, Matthias~De Lange, Marc Masana, Jary Pomponi, Gido
  van~de Ven, Martin Mundt, Qi~She, Keiland Cooper, Jeremy Forest, Eden
  Belouadah, Simone Calderara, German~I. Parisi, Fabio Cuzzolin, Andreas
  Tolias, Simone Scardapane, Luca Antiga, Subutai Amhad, Adrian Popescu,
  Christopher Kanan, Joost van~de Weijer, Tinne Tuytelaars, Davide Bacciu, and
  Davide Maltoni.
\newblock Avalanche: an end-to-end library for continual learning, 2021.

\bibitem[Lopez-Paz \& Ranzato(2017)Lopez-Paz and Ranzato]{lopez2017gradient}
David Lopez-Paz and Marc'Aurelio Ranzato.
\newblock Gradient episodic memory for continual learning.
\newblock In \emph{Advances in Neural Information Processing Systems (NIPS)},
  2017.

\bibitem[Mallya \& Lazebnik(2018)Mallya and Lazebnik]{mallya2018packnet}
Arun Mallya and Svetlana Lazebnik.
\newblock Packnet: Adding multiple tasks to a single network by iterative
  pruning, 2018.

\bibitem[Maurer et~al.(2016)Maurer, Pontil, and
  Romera-Paredes]{maurerBenefitMultitaskRepresentation}
Andreas Maurer, Massimiliano Pontil, and Bernardino Romera-Paredes.
\newblock The benefit of multitask representation learning.
\newblock \emph{J. Mach. Learn. Res.}, 17\penalty0 (1):\penalty0 2853–2884,
  January 2016.
\newblock ISSN 1532-4435.

\bibitem[Mendez et~al.(2020)Mendez, Wang, and Eaton]{mendez2020lifelong}
Jorge~A. Mendez, Boyu Wang, and Eric Eaton.
\newblock Lifelong policy gradient learning of factored policies for faster
  training without forgetting, 2020.

\bibitem[Mnih et~al.(2015)Mnih, Kavukcuoglu, Silver, Rusu, Veness, Bellemare,
  Graves, Riedmiller, Fidjeland, Ostrovski, et~al.]{mnih2015human}
Volodymyr Mnih, Koray Kavukcuoglu, David Silver, Andrei~A Rusu, Joel Veness,
  Marc~G Bellemare, Alex Graves, Martin Riedmiller, Andreas~K Fidjeland, Georg
  Ostrovski, et~al.
\newblock Human-level control through deep reinforcement learning.
\newblock \emph{nature}, 518\penalty0 (7540):\penalty0 529--533, 2015.

\bibitem[Mnih et~al.(2016)Mnih, Badia, Mirza, Graves, Lillicrap, Harley,
  Silver, and Kavukcuoglu]{mnih2016asynchronous}
Volodymyr Mnih, Adria~Puigdomenech Badia, Mehdi Mirza, Alex Graves, Timothy
  Lillicrap, Tim Harley, David Silver, and Koray Kavukcuoglu.
\newblock Asynchronous methods for deep reinforcement learning.
\newblock In \emph{International conference on machine learning}, pp.\
  1928--1937. PMLR, 2016.

\bibitem[Nguyen et~al.(2018)Nguyen, Li, Bui, and Turner]{VCL}
Cuong~V. Nguyen, Yingzhen Li, Thang~D. Bui, and Richard~E. Turner.
\newblock Variational continual learning.
\newblock In \emph{International Conference on Learning Representations
  (ICLR)}, 2018.

\bibitem[Parisi et~al.(2019)Parisi, Kemker, Part, Kanan, and
  Wermter]{parisi2019continual}
German~I Parisi, Ronald Kemker, Jose~L Part, Christopher Kanan, and Stefan
  Wermter.
\newblock Continual lifelong learning with neural networks: A review.
\newblock \emph{Neural Networks}, 2019.

\bibitem[Parisotto et~al.(2016)Parisotto, Ba, and
  Salakhutdinov]{parisotto16_actormimic}
Emilio Parisotto, Jimmy Ba, and Ruslan Salakhutdinov.
\newblock Actor-mimic: Deep multitask and transfer reinforcement learning.
\newblock In \emph{ICLR}, 2016.

\bibitem[Prabhu et~al.(2020)Prabhu, Torr, and Dokania]{prabhu12356gdumb}
Ameya Prabhu, Philip~HS Torr, and Puneet~K Dokania.
\newblock Gdumb: A simple approach that questions our progress in continual
  learning.
\newblock 2020.
\newblock URL
  \url{http://www.robots.ox.ac.uk/~tvg/publications/2020/gdumb.pdf}.

\bibitem[Raffin et~al.(2019)Raffin, Hill, Ernestus, Gleave, Kanervisto, and
  Dormann]{stable-baselines3}
Antonin Raffin, Ashley Hill, Maximilian Ernestus, Adam Gleave, Anssi
  Kanervisto, and Noah Dormann.
\newblock Stable baselines3.
\newblock \url{https://github.com/DLR-RM/stable-baselines3}, 2019.

\bibitem[Rebuffi et~al.(2017)Rebuffi, Kolesnikov, Sperl, and
  Lampert]{rebuffi2017icarl}
Sylvestre-Alvise Rebuffi, Alexander Kolesnikov, Georg Sperl, and Christoph~H
  Lampert.
\newblock icarl: Incremental classifier and representation learning.
\newblock In \emph{Computer Vision and Pattern Recognition (CVPR)}, 2017.

\bibitem[Riemer et~al.(2018)Riemer, Cases, Ajemian, Liu, Rish, Tu, and
  Tesauro]{riemer2018learning}
Matthew Riemer, Ignacio Cases, Robert Ajemian, Miao Liu, Irina Rish, Yuhai Tu,
  and Gerald Tesauro.
\newblock Learning to learn without forgetting by maximizing transfer and
  minimizing interference.
\newblock \emph{arXiv preprint arXiv:1810.11910}, 2018.

\bibitem[Ring(1997)]{ring1997child}
Mark~B Ring.
\newblock Child: A first step towards continual learning.
\newblock \emph{Machine Learning}, 28\penalty0 (1):\penalty0 77--104, 1997.

\bibitem[Rolnick et~al.(2019)Rolnick, Ahuja, Schwarz, Lillicrap, and
  Wayne]{rolnick2019experience}
David Rolnick, Arun Ahuja, Jonathan Schwarz, Timothy Lillicrap, and Gregory
  Wayne.
\newblock Experience replay for continual learning.
\newblock In \emph{Advances in Neural Information Processing Systems}, 2019.

\bibitem[{Rusu} et~al.(2016){Rusu}, {Rabinowitz}, {Desjardins}, {Soyer},
  {Kirkpatrick}, {Kavukcuoglu}, {Pascanu}, and {Hadsell}]{Rusu16progressive}
A.~A. {Rusu}, N.~C. {Rabinowitz}, G.~{Desjardins}, H.~{Soyer},
  J.~{Kirkpatrick}, K.~{Kavukcuoglu}, R.~{Pascanu}, and R.~{Hadsell}.
\newblock {Progressive Neural Networks}.
\newblock \emph{ArXiv e-prints}, 2016.

\bibitem[Schulman et~al.(2017)Schulman, Wolski, Dhariwal, Radford, and
  Klimov]{schulman2017proximal}
John Schulman, Filip Wolski, Prafulla Dhariwal, Alec Radford, and Oleg Klimov.
\newblock Proximal policy optimization algorithms.
\newblock \emph{arXiv preprint arXiv:1707.06347}, 2017.

\bibitem[Serr{\`{a}} et~al.(2018)Serr{\`{a}}, Sur{\'{\i}}s, Miron, and
  Karatzoglou]{HAT_Serra}
Joan Serr{\`{a}}, D{\'{\i}}dac Sur{\'{\i}}s, Marius Miron, and Alexandros
  Karatzoglou.
\newblock Overcoming catastrophic forgetting with hard attention to the task.
\newblock \emph{CoRR}, abs/1801.01423, 2018.

\bibitem[Shin et~al.(2017)Shin, Lee, Kim, and Kim]{Shin17}
Hanul Shin, Jung~Kwon Lee, Jaehong Kim, and Jiwon Kim.
\newblock Continual learning with deep generative replay.
\newblock In \emph{Advances in Neural Information Processing Systems (NIPS)},
  2017.

\bibitem[Tasfi(2016)]{tasfi2016PLE}
Norman Tasfi.
\newblock Pygame learning environment.
\newblock \url{https://github.com/ntasfi/PyGame-Learning-Environment}, 2016.

\bibitem[Taylor \& Stone(2009)Taylor and Stone]{JMLR09-taylor}
Matthew~E.\ Taylor and Peter Stone.
\newblock Transfer learning for reinforcement learning domains: A survey.
\newblock \emph{Journal of Machine Learning Research}, 10\penalty0
  (1):\penalty0 1633--1685, 2009.

\bibitem[Thrun \& Mitchell(1995)Thrun and Mitchell]{thrun1995lifelong}
Sebastian Thrun and Tom~M Mitchell.
\newblock Lifelong robot learning.
\newblock \emph{Robotics and autonomous systems}, 15\penalty0 (1-2):\penalty0
  25--46, 1995.

\bibitem[Thrun \& Schwartz(1995)Thrun and Schwartz]{thrun1995finding}
Sebastian Thrun and Anton Schwartz.
\newblock Finding structure in reinforcement learning.
\newblock In \emph{Advances in neural information processing systems}, pp.\
  385--392, 1995.

\bibitem[Traor{\'{e}} et~al.(2019)Traor{\'{e}}, Caselles{-}Dupr{\'{e}}, Lesort,
  Sun, Cai, Rodr{\'{\i}}guez, and Filliat]{Traore19DisCoRL}
Ren{\'{e}} Traor{\'{e}}, Hugo Caselles{-}Dupr{\'{e}}, Timoth{\'{e}}e Lesort,
  Te~Sun, Guanghang Cai, Natalia~D{\'{\i}}az Rodr{\'{\i}}guez, and David
  Filliat.
\newblock Discorl: Continual reinforcement learning via policy distillation.
\newblock \emph{CoRR}, abs/1907.05855, 2019.
\newblock URL \url{http://arxiv.org/abs/1907.05855}.

\bibitem[van~de Ven \& Tolias(2019)van~de Ven and Tolias]{van2019three}
Gido~M van~de Ven and Andreas~S Tolias.
\newblock Three scenarios for continual learning.
\newblock \emph{arXiv preprint arXiv:1904.07734}, 2019.

\bibitem[Veniat et~al.(2020)Veniat, Denoyer, and Ranzato]{ctrl}
Tom Veniat, Ludovic Denoyer, and Marc'Aurelio Ranzato.
\newblock Efficient continual learning with modular networks and task-driven
  priors, 2020.
\newblock URL \url{https://arxiv.org/abs/2012.12631}.

\bibitem[Wolczyk et~al.(2021)Wolczyk, Zajac, Pascanu, Kucinski, and
  Milos]{ContinualWorld}
Maciej Wolczyk, Michal Zajac, Razvan Pascanu, Lukasz Kucinski, and Piotr Milos.
\newblock Continual world: {A} robotic benchmark for continual reinforcement
  learning.
\newblock \emph{CoRR}, abs/2105.10919, 2021.
\newblock URL \url{https://arxiv.org/abs/2105.10919}.

\bibitem[Xie et~al.(2020)Xie, Harrison, and Finn]{xie2020deep}
Annie Xie, James Harrison, and Chelsea Finn.
\newblock Deep reinforcement learning amidst lifelong non-stationarity.
\newblock \emph{arXiv preprint arXiv:2006.10701}, 2020.

\bibitem[Yu et~al.(2019)Yu, Quillen, He, Julian, Hausman, Finn, and
  Levine]{MetaWorld}
Tianhe Yu, Deirdre Quillen, Zhanpeng He, Ryan Julian, Karol Hausman, Chelsea
  Finn, and Sergey Levine.
\newblock Meta-world: {A} benchmark and evaluation for multi-task and meta
  reinforcement learning.
\newblock \emph{CoRR}, abs/1910.10897, 2019.
\newblock URL \url{http://arxiv.org/abs/1910.10897}.

\bibitem[{Zenke} et~al.(2017){Zenke}, {Poole}, and {Ganguli}]{Zenke17}
Friedeman {Zenke}, Ben {Poole}, and Surya {Ganguli}.
\newblock Continual learning through synaptic intelligence.
\newblock In \emph{International Conference on Machine Learning (ICML)}, 2017.

\bibitem[Zeno et~al.(2019)Zeno, Golan, Hoffer, and Soudry]{zeno2019task}
Chen Zeno, Itay Golan, Elad Hoffer, and Daniel Soudry.
\newblock Task agnostic continual learning using online variational bayes,
  2019.

\end{thebibliography}
